\documentclass[sigconf,nonacm,10pt]{quantharness-preprint}

\usepackage{balance}
\usepackage{graphicx}
\usepackage{microtype}
\usepackage{placeins}
\usepackage{xurl}
\usepackage{tabularx}
\usepackage{longtable}
\usepackage{tikz}
\usetikzlibrary{arrows.meta,positioning}
\usepackage{tcolorbox}
\tcbuselibrary{breakable}
\usepackage{listings}

\newtcolorbox{trajectorybox}[2]{%
  breakable,title={#1\quad #2},
  title after break={#1\quad #2\space(continued)},
  colback=white,colframe=black!65,colbacktitle=black!6,coltitle=black,
  fonttitle=\small\bfseries,fontupper=\small\raggedright,
  boxrule=0.4pt,arc=1mm,left=2mm,right=2mm,top=1.5mm,bottom=1.5mm,
  before skip=7pt,after skip=4pt,
}
\newtcolorbox{promptbox}[1]{%
  breakable,title={#1},title after break={#1\space(continued)},
  colback=blue!2!white,colframe=blue!40!black,
  colbacktitle=blue!9!white,coltitle=black,
  fonttitle=\small\bfseries,fontupper=\small\raggedright,
  boxrule=0.6pt,arc=1mm,left=2mm,right=2mm,top=1.5mm,bottom=1.5mm,
  before skip=8pt,after skip=6pt,
  before upper={\setlength{\parskip}{4pt}},
}
\newenvironment{promptitems}{%
  \par\begingroup
  \setlength{\parskip}{2pt}%
  \interlinepenalty=10000\relax
  \renewcommand{\item}{%
    \par\noindent\hangindent=1.3em\hangafter=1\relax
    \textbullet\hspace{0.5em}\ignorespaces}%
}{\par\endgroup}
\lstdefinestyle{trajectorypython}{%
  language=Python,basicstyle=\ttfamily\fontsize{10}{12}\selectfont,
  keywordstyle=\bfseries,commentstyle=\itshape,
  columns=fullflexible,keepspaces=true,showstringspaces=false,
  breaklines=true,breakindent=1em,aboveskip=3pt,belowskip=3pt,
}

\renewcommand{\small}{\fontsize{10}{12}\selectfont}
\renewcommand{\footnotesize}{\fontsize{10}{12}\selectfont}
\renewcommand{\scriptsize}{\fontsize{10}{12}\selectfont}

\graphicspath{{figures/}}

\setcopyright{none}
\acmConference[]{}{}{}{}
\acmBooktitle{}
\copyrightyear{}
\acmYear{2026}
\acmDOI{}
\acmISBN{}
\acmSubmissionID{}
\submissionType{}

\newcommand{\preprintpagestyle}{%
  \fancyhf{}%
  \fancyfoot[C]{\small\thepage}%
  \renewcommand{\headrulewidth}{0pt}%
  \renewcommand{\footrulewidth}{0pt}%
}
\AtBeginDocument{%
  \fancypagestyle{standardpagestyle}{\preprintpagestyle}%
  \fancypagestyle{firstpagestyle}{\preprintpagestyle}%
  \pagestyle{standardpagestyle}%
}
\hypersetup{hidelinks,bookmarksnumbered=true}

\title[QuantHarness for Financial Strategy Development]{Reliable Before Profitable: QuantHarness for Governing Financial Strategy Development with LLM Agents}

\author{Jiasheng Mao}
\authornote{These authors contributed equally to this work.}
\affiliation{
  \institution{ShanghaiTech University}
  \city{Shanghai}
  \country{China}}
\email{maojsh2025@shanghaitech.edu.cn}

\author{Jiacheng Liu}
\authornotemark[1]
\affiliation{
  \institution{ShanghaiTech University}
  \city{Shanghai}
  \country{China}}
\email{liujch12025@shanghaitech.edu.cn}

\author{Hengwei Ye}
\affiliation{
  \institution{ShanghaiTech University}
  \city{Shanghai}
  \country{China}}
\email{yehw2024@shanghaitech.edu.cn}

\author{Zheng Tian}
\authornote{Corresponding author.}
\affiliation{
  \institution{ShanghaiTech University}
  \city{Shanghai}
  \country{China}}
\email{tianzheng@shanghaitech.edu.cn}

\renewcommand{\shortauthors}{Mao et al.}

\begin{abstract}
LLM agents can turn natural-language financial research requests into executable
and backtested strategies, but a successful run may still implement the wrong
rule, use information unavailable at decision time, or obscure how repairs and
candidate selection changed the strategy. Because these failures arise
throughout development, final code and performance metrics alone are
insufficient to establish either fidelity to the request or the validity of the
development process. We therefore address the reliability of financial strategy 
development at the trajectory level: the system must control how intermediate outputs
become downstream state and preserve the evidence needed to justify the terminal
outcome. We introduce QuantHarness, a two-layer agent harness in which a
finance-specific Workflow structures specification, data grounding,
construction, testing, and review, while Harness Core enforces stage-specific
information, action, and commitment boundaries. We also introduce FinRelBench, 
a benchmark that evaluates strategy-development reliability through execution success, 
requirement satisfaction, and runtime evidence. QuantHarness achieves the highest Task Success 
at 90.72\%, compared with 74.23\%--84.54\% for the comparison systems. Controlled stress tests show that 
its boundaries contain targeted violations, while structured run records improve evidence retrieval and support decision replay under the evaluated protocol.
\end{abstract}


\begin{document}

\raggedbottom
\maketitle

\section{Introduction}
\label{sec:introduction}

Recent LLM-based systems can carry quantitative strategy development from a
natural-language research request to an implemented and backtested strategy
\citep{li2025rdagentquant,zhang2026alphaforgebench}.
Yet successful execution alone does not establish that the resulting strategy
is faithful to the original request. Generated code may implement a different
trading rule from the one requested \citep{khoroshilov2026quantcodebench}, use
information unavailable at decision time, or introduce temporal misalignment
\citep{luo2026alphaqtbench}. Problems may also emerge after the initial
implementation has been produced. A repair can remove an execution error by changing the trading
rule itself. Repeated selection based on backtest results can turn
evaluation into search over historical outcomes
\citep{white2000realitycheck,bailey2017backtestoverfitting}. Because these
problems can arise at different stages, the final code and performance
metrics alone may not reveal their source or the decisions that allowed them to
persist. We call one execution of strategy development a \emph{run}. The ordered path
from its research request to its terminal outcome is the run's
\emph{development trajectory}. We therefore examine reliability across this trajectory: 
whether the requested strategy is faithfully implemented and the development process is supported by evidence that justifies its terminal outcome.

Prior work approaches parts of this problem from two directions. Financial
agents and benchmarks support trading decisions, factor discovery, executable
research, and strategy evaluation
\citep{xiao2025tradingagents,li2025rdagentquant,duan2025factormad,khoroshilov2026quantcodebench,zhang2026alphaforgebench,wang2026backtestbench},
whereas general agent-harness research studies how context, tools, persistent
state, and runtime control shape agent behavior
\citep{yang2024sweagent,mei2024aios,wang2026agentspec}. These directions 
motivate the question of how an LLM agent can reliably carry a financial strategy request 
through specification, data grounding, implementation, testing, and final acceptance.

We introduce \textsc{QuantHarness}, a two-layer system for governing this
trajectory.
The finance-specific \emph{Workflow} organizes the progression from
an explicit account of the intended strategy through implementation and testing
to a terminal outcome. \emph{Harness Core} mediates that progression by
controlling the information and actions available at each stage and deciding
which outputs may update the run state used downstream. Under these controls, 
the run ends with either a frozen strategy or termination without acceptance. 
QuantHarness also preserves a structured \emph{run record} that links the run's 
terminal outcome to its supporting artifacts and lifecycle decisions.

We evaluate trajectory reliability on FinRelBench, a benchmark of 97 explicit rule-based tasks. 
Our evaluation examines end-to-end Task Success, boundary enforcement, post-run audit and replay. 
The results show that QuantHarness achieves the highest Task Success among the compared systems, contains targeted violations in the stress tests, and supports more effective evidence retrieval and decision replay through structured run records.
Reliable development provides a basis for interpreting financial outcomes, but
a financial research system should also support economically meaningful
strategy outputs. We therefore use FinResearchSuite as an auxiliary evaluation
of QuantHarness's practical utility for financial research, with results
reported in Appendix~\ref{app:e3-results}. Our contributions are as follows:
\begin{itemize}
    \item We frame reliable financial strategy development as a problem of
    trajectory governance and distinguish trajectory reliability from financial
    quality.
    \item We introduce QuantHarness, a two-layer system that couples a
    finance-specific Workflow with Harness Core to govern workflow execution
    and preserve an artifact-backed run record.
    \item We develop FinRelBench, a 97-task benchmark with a controlled evaluation 
    protocol for strategy-development reliability, and use its tasks to evaluate end-to-end 
    Task Success, boundary enforcement, post-run audit and replay.
\end{itemize}

\section{Related Work}
\label{sec:related-work}

\subsection{LLM Agents for Quantitative Finance}
\label{subsec:rw-financial-strategy-agents}

Recent work on LLM-based financial agents spans market-facing decision support
and executable quantitative research. Decision-oriented systems organize market
information through mechanisms such as memory, reflection, and role-based collaboration
\citep{yu2024finmem,xiao2025tradingagents,zhang2024finagent,yu2024fincon},
while platforms such as FinRobot~\citep{yang2024finrobot} support a broader range of financial-analysis
tasks. For executable research, Qlib~\citep{yang2020qlib} and
FinRL-Meta~\citep{liu2022finrlmeta} provide reusable data and market-simulation infrastructure.
R\&D-Agent-Quant~\citep{li2025rdagentquant} combines structured
specifications with implementation, backtesting, iterative feedback, and
retained research artifacts, while FactorMAD~\citep{duan2025factormad} uses agent debate for
interpretable alpha-factor discovery. As these systems automate more of
the research process, reliability concerns extend beyond the performance of the
resulting strategy to the decisions and evidence accumulated during its
development. Successful execution and financial results alone do not establish
that a request was appropriately interpreted, grounded in data, and carried
through to a justified outcome. QuantHarness addresses this broader need for
task fidelity, process control, and supporting evidence across the full
development trajectory.

\subsection{Evaluation of Financial Agents}
\label{subsec:rw-financial-agent-evaluation}

Recent evaluation efforts for financial LLM systems cover both broad financial
capabilities and executable research. FinBen~\citep{xie2024finben} illustrates this breadth, covering
financial language tasks, forecasting, risk management, and trading.
More specialized evaluations address the outputs of
quantitative research: AlphaForgeBench~\citep{zhang2026alphaforgebench} evaluates generated factors and
strategies through standardized backtesting, while BacktestBench~\citep{wang2026backtestbench} compares
automated backtesting outputs with reference answers.
At the code level, QuantCode-Bench~\citep{khoroshilov2026quantcodebench} and
AlphaQT-Bench~\citep{luo2026alphaqtbench} assess properties such as
alignment with task descriptions, temporal causality, functional correctness,
and structural compliance. Together, these benchmarks assess financial capabilities 
and the quality of research outputs. However, output-level assessment alone does not 
establish whether task constraints were respected throughout development. We therefore 
use FinRelBench to assess whether the requested strategy is faithfully implemented and 
its required behavior is demonstrated in execution. We also examine how agents reached that result, 
asking whether development respected task constraints and left sufficient evidence for subsequent audit.

\subsection{Agent Harnesses and Runtime Governance}
\label{subsec:rw-agent-harnesses}

Beyond the capabilities of individual models, recent work on LLM agents
examines the surrounding harness: the supporting execution environment that
organizes context, tool access, state, and workflow execution.
Frameworks such as AutoGen~\citep{wu2024autogen},
SWE-agent~\citep{yang2024sweagent},
OpenHands~\citep{wang2025openhands}, and
AIOS~\citep{mei2024aios} provide different combinations of these capabilities.
Recent studies also examine how to represent and improve the harness itself.
Natural-Language Agent Harnesses~\citep{pan2026naturalharnesses}
expresses control policies as editable documents executed through a shared
runtime, with explicit handoffs, state updates, and validation gates.
Meta-Harness~\citep{lee2026metaharness} optimizes harness code by inspecting
earlier candidates, evaluation scores, and execution traces.
Alongside this work on harness design,
AgentSpec~\citep{wang2026agentspec} enforces configurable rules during
agent execution. Together, these approaches provide mechanisms for
organizing, improving, and constraining agent behavior.
For financial strategy development, these mechanisms must be grounded
in trading requirements and market-data constraints.
The challenge is to carry those requirements through construction,
testing, and acceptance while retaining the evidence needed to justify
each outcome. QuantHarness investigates this problem across the full
development trajectory.

\newcommand{\methodlead}[1]{%
  \par\indent\textit{#1.}\enspace
}

\section{QuantHarness}
\label{sec:method}

\subsection{System Overview}
\label{subsec:system-overview}

\begin{figure*}[t]
  \centering
  \includegraphics[width=0.95\textwidth]{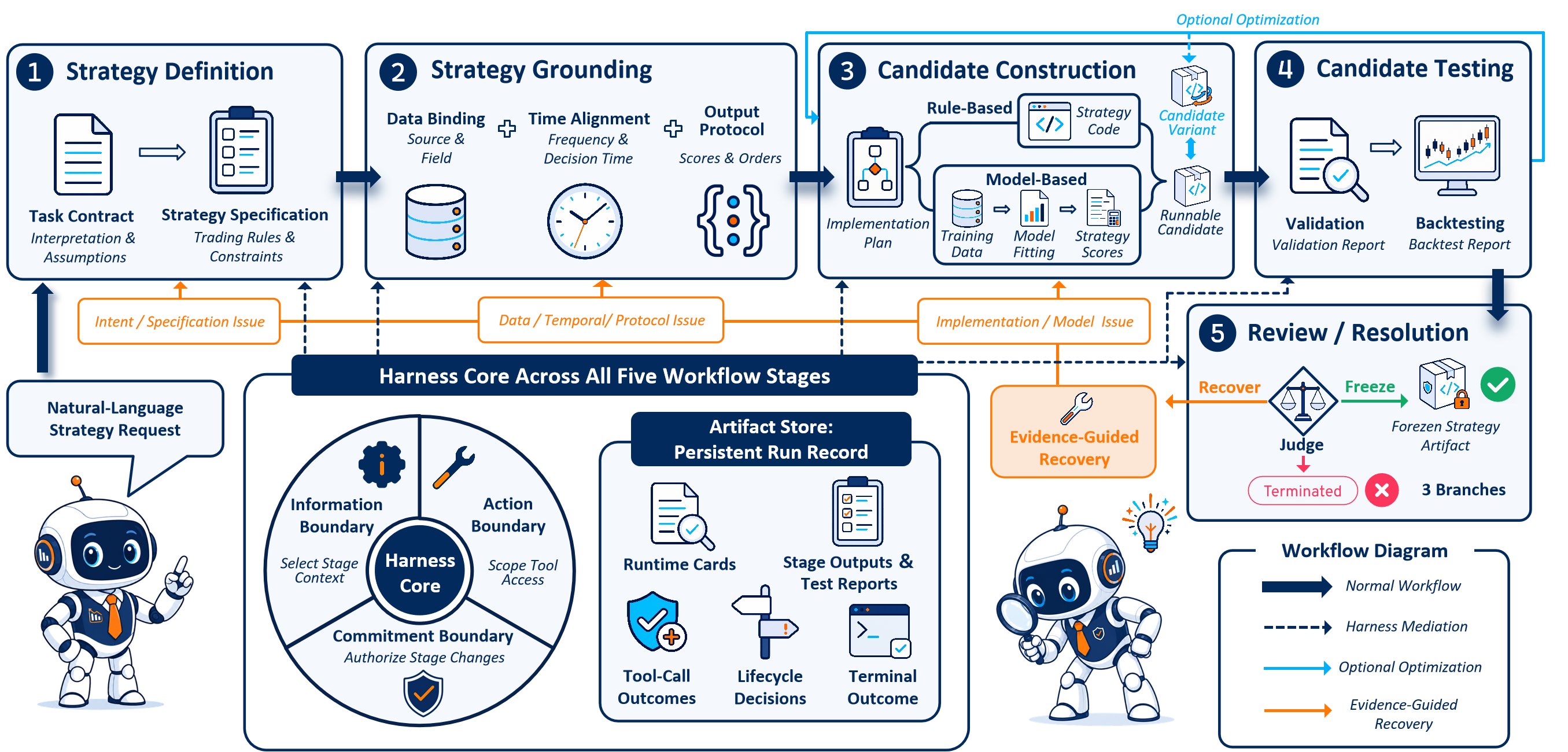}
  \caption{Overview: Workflow structures strategy development; Harness Core
  governs each run and preserves its run record.}
  \label{fig:quantharness-workflow}
  \Description{A two-layer overview of QuantHarness. The finance-specific
  Workflow progresses from strategy definition and grounding through candidate
  construction, testing, review, and resolution. Harness Core mediates
  information, actions, and lifecycle commitments and preserves an
  artifact-backed run record across these stages.}
\end{figure*}

Turning a natural-language request into a runnable financial strategy requires
both an explicit development process and control over how intermediate results
are used. QuantHarness governs the development trajectory through two
interacting layers: an agent-facing Workflow organizes strategy development,
while Harness Core governs its execution and preserves the resulting run
record.
Figure~\ref{fig:quantharness-workflow} summarizes how the two layers interact.
The following subsections describe the Workflow and Harness Core in turn.

\subsection{Finance-Specific Workflow}
\label{subsec:workflow}

Within QuantHarness, the Workflow spans the process from
clarifying a request to producing and testing a runnable strategy
implementation. The remainder of this subsection presents the stages of this
process in sequence.

\methodlead{Making the requested strategy explicit}
A natural-language strategy request often leaves information required for
implementation unspecified. If the model fills these gaps only while writing
code, its assumptions become part of the implementation without being stated
explicitly. Later stages then cannot compare the code with an explicit
description of the intended strategy. QuantHarness therefore first documents its
interpretation of the request, including any assumptions and unresolved
information. The resulting structured description is the \emph{task contract}. The
Workflow then uses the contract to produce a \emph{strategy specification}.
Whereas the task contract captures what the system understood from the request,
the strategy specification states the trading rules and constraints that the
eventual implementation must satisfy. The specification is the common
reference for connecting the strategy to available data and implementing it in
code. Any later change in strategy meaning must first be reflected in the
specification.

\methodlead{Grounding strategy requirements in available data and output formats}
A strategy specification identifies the data a trading rule needs, but it does
not establish whether a compatible source and field exist among the available
datasets. Deferring this check to code generation may cause the implementation
to refer to nonexistent fields or use data at an incompatible frequency.
QuantHarness therefore inspects the available data before implementation rather
than assuming that every required input exists. A \emph{data binding} maps each
requirement to a confirmed data source, field, and frequency; requirements that
cannot be resolved remain explicit rather than being bound to guessed inputs.
A data binding alone is not sufficient for testing because the backtester must
also know how to interpret the strategy's decisions. An \emph{output protocol}
specifies whether these decisions are represented as asset scores or trade
orders and defines the fields expected by the backtester. Together, the data
binding and output protocol define how the strategy receives data and
communicates decisions without changing the trading rule itself.

\methodlead{Constructing a runnable strategy implementation}
The preceding stages fix the strategy requirements, data inputs, and output
protocol, but they do not yet specify the computation that will produce the
trading decisions. If this procedure is determined only during code generation
or model training, the choices that define it remain implicit in the resulting
implementation. QuantHarness therefore makes these choices explicit in an
\emph{implementation plan} before
construction begins. The plan states how the bound inputs will be transformed
into trading decisions under the strategy specification and output protocol.
QuantHarness realizes this plan through one of two supported paths. Rule-based
strategies are implemented as code that follows the specified calculation.
Model-based strategies use a separate construction path. To keep model
construction within a testable interface, the current path is limited to a
predefined library of machine-learning and deep-learning templates provided by
QuantHarness. The system first constructs training data from the bound inputs
and fits a model instantiated from the selected template. The fitted model then
produces strategy scores in the declared output format. The resulting strategy
implementation, whether rule-based or model-based, becomes the candidate
examined in the next stage.

\methodlead{Testing the candidate against established requirements}
The implementation plan guides construction, but the resulting candidate may
still fail to satisfy the structural, data, or execution requirements
established earlier. QuantHarness therefore tests the completed candidate
against those requirements before it can proceed. The testing stage has two
required steps: (1) \emph{validation} and (2) \emph{backtesting}.
Validation first checks whether the candidate is structurally complete and
compatible with the execution environment. It then checks whether the candidate
uses the declared inputs and provides trading decisions in the required form.
The results, including any failed checks, are listed in a validation report.
Only candidates that pass validation proceed to backtesting.
Backtesting applies the candidate to the market data specified by the data
binding and passes its trading decisions to the backtester under the declared
output protocol. The resulting backtest report indicates whether backtest
execution succeeds and whether the trading decisions can be processed by the
backtester. When backtest execution succeeds, the report also includes the
resulting financial performance metrics. These financial values do not override the validation
and execution checks. A candidate proceeds only after both checks pass.

\methodlead{Optional strategy optimization}
Passing validation and backtesting establishes that a candidate can be executed
with the bound inputs and declared output protocol. At this point, QuantHarness
supports an optional, budget-limited exploration of whether its financial quality can be
improved without changing the task requirements, data binding, or output
protocol. Within this loop, candidate variants must pass validation before
being backtested on the optimization period and compared under a preset
objective. The selected candidate is then evaluated once on a separate final
period. This result is not returned to the loop, and its financial values
support quality analysis rather than determine engineering acceptance.
The candidate then proceeds to Judge review with the corresponding validation
and backtest reports. The full optimization protocol is provided in
Appendix~\ref{app:strategy-optimization}.

\methodlead{Candidate review and Workflow resolution}
After candidate development and testing are complete, the Workflow must decide
whether the candidate can be accepted or requires further action. QuantHarness
assigns this review to the \emph{Judge}, which considers the candidate together
with the artifacts and reports produced earlier in the Workflow. The review
leads to three routes: recovery, termination, or freezing. When the Judge identifies a
correctable problem, its report states which earlier stage needs to be
revisited. Recovery returns the run to that stage, after which the affected
downstream stages are repeated within a fixed attempt budget. If the problem
cannot be resolved or the budget is exhausted, the run terminates without an
accepted strategy. If the available evidence supports acceptance, freezing
creates a \emph{frozen strategy artifact} that preserves the candidate
implementation and the evidence supporting its acceptance. Harness Core checks
the evidence supporting the Judge's recommendation before it allows the
corresponding change to take effect.

\subsection{Harness Core: Runtime Mediation and Governance}
\label{subsec:harness-core}

The Workflow organizes strategy development into stages, but stage order alone
does not control how each stage operates or how its results affect later
stages. QuantHarness therefore introduces Harness Core, a runtime control layer
that mediates stage execution across the Workflow. The \emph{information
boundary} determines which existing run
information is supplied to a stage. A separate \emph{action boundary} limits
the external capabilities that the stage may invoke. Once a stage produces an
output, the \emph{commitment boundary} determines whether later stages may rely
on it. Across these three boundaries, Harness Core preserves the artifacts and
lifecycle decisions needed to explain how the run progressed.

\methodlead{Providing stage-specific context}
Each Workflow stage requires information tailored to its current task. The full
history mixes current artifacts with superseded candidates, evidence from
earlier failures, and control or audit metadata. Because the relevance of this
material changes across stages, supplying the full history would leave the
agent to determine what remains valid. Harness Core instead constructs a
distinct \emph{Runtime Card} for each agent-facing stage. Its agent-visible
portion states the stage goal, required outputs, and applicable constraints,
together with the task information and feedback relevant at that point in the
run. Control and audit information remains private to Harness Core. This
stage-specific selection gives the agent a defined context without requiring it
to interpret the full run history.

\methodlead{Mediating tools and external effects}
Not every external capability is appropriate at every Workflow stage.
Unrestricted access could allow a stage to trigger an operation intended for
another part of the Workflow or to act before its required inputs exist.
QuantHarness enforces the action boundary by representing each
agent-accessible capability as a \emph{tool} mediated by Harness Core. A
\emph{tool registry} gives every tool a stable name and a declared input
structure. For each stage, Harness Core exposes only the registered tools
allowed for its current task. Calls to these tools are checked against their
declared input structures and routed through Harness Core to the corresponding
capabilities. Tool access is therefore enforced by Harness Core rather than
left to prompt instructions.

\methodlead{Authorizing and committing lifecycle changes}
A Workflow stage may recommend what should happen next.
Allowing this recommendation to change the run could advance the Workflow
before the required outputs and evidence have been checked. QuantHarness
separates the stage's proposal from the authorization and state update performed
by Harness Core. The run's \emph{lifecycle} is its progression through Workflow
stages from creation to a terminal result. Before a proposed transition changes
this lifecycle, Harness Core intercepts it through a \emph{lifecycle hook}. A
\emph{guard} then checks whether the proposed route is allowed and whether the
artifacts and evidence required by the target stage are available.
The guard approves or rejects the proposal without altering the run.
On approval, the \emph{runtime} commits the transition by updating the current
stage and making the outputs available downstream.

\methodlead{Preserving run evidence and artifacts}
A final strategy implementation and its backtest result do not explain how a
run reached its outcome. QuantHarness therefore maintains an \emph{Artifact
Store} for each run. During the run, the store retains the Runtime Cards
supplied to agent-facing stages, tool-call outcomes, lifecycle decisions, and
the outputs produced at each stage, including outputs that later fail a check.
When the Workflow freezes a candidate, the store links the final implementation
to the intermediate artifacts and decisions supporting its acceptance. If no
candidate is accepted, the store retains the evidence showing where and
why the run ended. Post-run analyses can inspect these materials but cannot
modify them or feed new measurements back to the Workflow. The Artifact Store
thus preserves the terminal outcome together with
the intermediate evidence needed to inspect how it was produced.

\section{FinRelBench}
\label{sec:benchmark-design}

\subsection{Benchmark Objective}
\label{subsec:finrelbench-objective}

Reliable financial strategy development requires an agent to preserve the
user's requirements throughout requirement interpretation, data preparation,
strategy implementation, runtime verification, and final acceptance. A strategy
may run successfully and produce favorable backtest results even when its data
processing or trading logic departs from those requirements. Evaluating this
reliability therefore requires tasks that connect explicit strategy requirements
to verifiable implementation and execution evidence. To meet this need, we
construct \textbf{FinRelBench} (Financial Strategy Development Reliability
Benchmark) to assess whether agents translate stated requirements
into correct data processing, strategy implementations, and observed runtime
behavior.

FinRelBench assesses this translation at three complementary levels:
\emph{executability}, whether the data pipeline and strategy run successfully;
\emph{requirement satisfaction}, whether the implementation meets the stated
data and trading requirements; and \emph{runtime evidence}, whether execution
records demonstrate the required behavior. Together, these levels support
end-to-end evaluation of task completion. The same tasks also provide common
workloads for boundary-control experiments and audit and replay studies, which
examine how errors are controlled during development and how outcomes can be
traced and reproduced. 

\subsection{Benchmark Composition}
\label{subsec:finrelbench-composition}

FinRelBench comprises 97 explicitly specified, rule-based strategy-development
tasks in the A-share equity setting: 80 tasks selected and adapted from the
400-task QuantCode-Bench~\citep{khoroshilov2026quantcodebench}, and 17 tasks of
our own design. We select source tasks whose core trading logic can be expressed
as verifiable requirements using the available market data under long-only
equity trading. Tasks whose defining requirements depend on short selling,
derivatives, or unavailable external data are excluded. Adaptation preserves
the core trading logic while aligning the stock universe, market data, and
execution timing with A-share trading rules, including T+1 settlement.

The 80 adapted tasks cover trend and momentum, breakouts, mean reversion,
volatility and liquidity conditions, and combinations of indicators, including
rules that depend on position state or calendar schedules. Under QuantCode-Bench's
original difficulty labels, these 80 tasks comprise 41 easy, 29 medium, and
10 hard cases. The 17 tasks of our own design test data adjustment and temporal
alignment, stock eligibility and
portfolio constraints, cash fallback, and execution timing. Together, these
tasks test whether agents preserve both the specified trading logic and the
data and execution requirements needed to implement it correctly.

The combined suite contains 64 daily-score tasks (50 adapted and 14 original)
and 33 minute-order tasks (30 adapted and three original). Daily-score tasks assign scores to
stocks for cross-sectional ranking and portfolio construction, testing whether
agents correctly implement historical windows, ranking rules, and portfolio
constraints. Minute-order tasks produce timestamped buy and sell instructions,
testing bar aggregation, decision windows, order delays, and position-state
transitions. These output types provide complementary checks of how strategy
requirements are translated into data processing and actual trading behavior.
Task examples are provided in Appendix~\ref{app:benchmark-details}.

FinRelBench evaluates reliability against explicit strategy requirements.
While reliable development provides a basis for interpreting financial
outcomes, practical use in financial research also requires evidence of
economically meaningful strategy outputs.
To complement this primary evaluation with evidence of practical financial
utility, we introduce
\textbf{FinResearchSuite} (Financial Strategy Development Research Suite).
This auxiliary suite contains 22 tasks: 12 open-ended rule-research tasks,
5 specified-model development tasks, and 5 open-ended model-research
tasks. It assesses research completion and the return and risk of the
resulting strategies to examine whether QuantHarness's governed development
process can support economically meaningful strategy research. Financial
performance thus serves as an auxiliary check of basic research utility and
does not enter FinRelBench's reliability scores. Task definitions and results
are provided in Appendices~\ref{app:finresearchsuite} and~\ref{app:e3-results}.

\subsection{Evaluation Protocol}
\label{subsec:finrelbench-protocol}

Each FinRelBench case is a self-contained evaluation package divided into agent-visible and evaluator-only components. 
The agent-visible components are the \emph{public task query}, \emph{market snapshot}, 
and \emph{data and submission documentation}. 
The evaluator-only components are the \emph{trade execution settings} and \emph{task-specific rubric}. 
The \emph{public task query} states the task objective and public constraints. The \emph{market snapshot} provides versioned, read-only market data, 
while the \emph{data and submission documentation} describes the tables, column definitions, time coverage, and access paths, explaining how to access the data and submit results.
The \emph{trade execution settings} specify the test period, trade execution rules,
and transaction costs. The \emph{task-specific rubric} defines the requirements
used to assess the implementation and its execution results. 
Because different implementations can correctly fulfill the same request, these
requirements are expressed as checkable criteria. 
Data criteria examine whether the requested securities
and fields are used, enough past observations are available for the specified
calculations, and the prepared data are actually consumed. Strategy criteria
examine the requested signal calculation, stock selection, and trading
schedule. 
Criteria labeled \emph{must} must all be satisfied for minimum
correctness, while partial satisfaction of supporting \emph{should} criteria
allows some implementation flexibility. 
The criteria describe the expected data processing and strategy behavior, while forbidden
conditions identify behavior that invalidates the result regardless of
other requirements satisfied. For example, a strategy may implement the
requested trading rule but use future prices to generate earlier signals.
Confirmed forbidden behavior therefore causes task failure, even when
the implementation passes execution checks and satisfies other criteria.

Deterministic mechanical checks establish whether the agent's code can
construct the data and execute the strategy needed for task completion.
Systems provide their code, data-construction and strategy-execution
commands, and saved data and outputs through a common interface.
The evaluator runs these commands under fixed market and execution
conditions. \emph{Data Pipeline Success} requires the dataset to be
constructed, saved, and consumed by the strategy; \emph{Strategy Run Success}
requires the strategy to complete execution and produce its declared output.

A strategy can run successfully while using incorrect data or trading rules,
so an external panel uses the rubric to check whether the implementation
meets the task requirements. The panel examines anonymized code, summaries of the prepared
datasets, execution records, and outputs. For each criterion, it separately
assesses whether the requirement is satisfied and whether execution evidence
demonstrates the relevant behavior. Runtime support requires both code on
the actual execution path and corresponding evidence from the run or its
outputs. Independent evaluators assess two equivalent wordings of the same
rubric. Disagreements that can affect the result are resolved by an arbiter
who reviews the evidence without seeing the earlier judgments.

These assessments jointly determine \emph{Task Success}. Both mechanical
checks must pass, and each semantic dimension must satisfy all must criteria,
at least 75\% of should criteria rounded up, and no confirmed forbidden
condition or independently confirmed fatal defect. Additionally, every
criterion derived from an explicit public requirement must be satisfied and
supported by runtime evidence, including those marked should.

Alongside the two mechanical metrics, we report \emph{Requirement Satisfaction}
and \emph{Runtime-Evidence Coverage} separately for data and strategy,
giving six diagnostics in total. These criterion-based metrics measure the
fractions judged satisfied and supported by execution evidence, respectively,
with equal weight for must and should criteria.
Section~\ref{subsec:schema-neutral-construction} specifies how task-level
outcomes and criterion-level judgments are aggregated for the reported
comparison. Appendix~\ref{app:benchmark-details} provides the detailed rubric,
evidence rules, and worked examples.

\section{Experiments}
\label{sec:experiments}

\subsection{Experimental Setup}
\label{subsec:experimental-setup}

Our evaluation is designed to establish three complementary forms of
reliability evidence. We first test whether QuantHarness, as a complete system, can
construct executable strategies that use the required data and realize the
requested behavior. We then examine how individual Harness Core boundaries
affect reliable completion and violation containment, followed by an evaluation
of whether structured run records support post-run audit and decision replay.
FinResearchSuite
separately assesses completion and financial utility in open-ended rule and
model research.

For strategy construction and research, we compare QuantHarness with Codex CLI~\cite{openai2025codex},
OpenHands~\cite{wang2025openhands}, SWE-agent~\cite{yang2024sweagent},
and Aider~\cite{gauthier2026aider}. All strategy-generation
calls use GPT-5.5~\cite{openai2026gpt55}. Within each task, these systems
receive the same public request, frozen market data, and documentation,
while retaining their native agent loops, tools, and recovery policies.

To compare systems with different internal workflows, FinRelBench evaluates
task completion using a common set of materials. Each system provides its
code, commands for preparing data and running the strategy, and saved data
and outputs. These materials allow the evaluator to apply the mechanical
and semantic checks defined in Section~\ref{subsec:finrelbench-protocol},
without requiring systems to reproduce QuantHarness's workflow stages
or internal record formats. 
Within each comparison,
a deterministic A-share backtester uses shared signal timing, transaction
costs, and trading constraints. Software versions, resource limits, data
windows, and execution parameters are specified in
Appendices~\ref{app:implementation} and~\ref{app:finresearchsuite-protocol}.

For FinRelBench semantic assessment, Claude Opus
4.6~\cite{anthropic2026claudeopus46} and Gemini 3.1
Pro~\cite{googledeepmind2026gemini31pro} independently examine anonymized
evidence under two equivalent rubric wordings. DeepSeek V4
Pro~\cite{deepseek2026v4pro} blindly adjudicates disagreements that can
change the outcome. These external assessments provide no feedback for
another development attempt. Appendix~\ref{app:benchmark-details} details the panel
procedure, and Appendix~\ref{app:e1-results} documents the implementation and evidence-collection
configurations represented in the reported comparison.

\subsection{Strategy Development Reliability}
\label{subsec:schema-neutral-construction}

A strategy may execute successfully while using unsuitable data, implementing
a different trading rule, or leaving required behavior unverified. We evaluate
this reliability gap on all 97 FinRelBench tasks, asking whether each system
carries an explicit request through data preparation and strategy implementation
to a verifiable final result. Task Success is the primary measure:
within the same task, both mechanical checks and minimum semantic gates
must pass, and every explicit public requirement must be satisfied and
supported by runtime evidence, as defined in
Section~\ref{subsec:finrelbench-protocol}.

Task Success and the two mechanical metrics are pass rates over all
97 planned tasks. Requirement Satisfaction and Runtime-Evidence Coverage
measure the fractions of criteria satisfied and supported by execution
evidence within each task and dimension. These fractions are averaged
across tasks, giving each task equal weight regardless of how many criteria
its rubric contains. Completed diagnostic assessments retain their scores
even when Task Success is zero, showing which requirements were fulfilled
within otherwise unsuccessful tasks. 
Incomplete assessments are distinguished from confirmed failures in Appendix~\ref{app:e1-results}.

\begin{table*}[t]
  \caption{Strategy-development reliability on 97 planned FinRelBench
  tasks per system. Higher is better.}
  \label{tab:schema-neutral-construction-results}
  \centering
  \scriptsize
  \setlength{\tabcolsep}{2.5pt}
  \renewcommand{\arraystretch}{1.08}
  \begin{tabular*}{\textwidth}{@{\extracolsep{\fill}}lccccccc@{}}
    \toprule
    & \multicolumn{1}{c}{\shortstack{Main\\measure}}
    & \multicolumn{2}{c}{Mechanical diagnostics}
    & \multicolumn{4}{c}{Requirement diagnostics} \\
    \cmidrule(lr){2-2}\cmidrule(lr){3-4}\cmidrule(l){5-8}
    System
      & \shortstack{Task\\Success}
      & \shortstack{Data Pipeline\\Success}
      & \shortstack{Strategy Run\\Success}
      & \shortstack{Data\\Requirement\\Satisfaction}
      & \shortstack{Data\\Runtime-\\Evidence\\Coverage}
      & \shortstack{Strategy\\Requirement\\Satisfaction}
      & \shortstack{Strategy\\Runtime-\\Evidence\\Coverage} \\
    \midrule
    QuantHarness
      & \textbf{90.72\%}
      & \textbf{100.00\%} & \textbf{98.97\%}
      & \textbf{98.97\%} & 98.28\%
      & 98.71\% & \textbf{96.07\%} \\
    Codex CLI
      & 81.44\%
      & 96.91\% & 91.75\%
      & 97.59\% & 96.56\%
      & 97.50\% & 91.11\% \\
    OpenHands
      & 74.23\%
      & 86.60\% & 84.54\%
      & 84.54\% & 84.54\%
      & 85.38\% & 81.40\% \\
    SWE-agent
      & 84.54\%
      & 94.85\% & 93.81\%
      & 94.85\% & 94.50\%
      & 94.44\% & 92.85\% \\
    Aider
      & 79.38\%
      & 98.97\% & 94.85\%
      & 96.91\% & \textbf{99.31\%}
      & \textbf{99.09\%} & 92.64\% \\
    \bottomrule
  \end{tabular*}
  \vspace{1pt}
  \parbox{\textwidth}{\scriptsize
    \textit{Notes.}
    Bold indicates the highest value in each column.
    Task Success and mechanical diagnostics are task pass rates.
    Requirement diagnostics are task-macro averages of criterion-level
    fractions, not the percentages of tasks satisfying all criteria.
    Every metric uses all 97 planned tasks; incomplete assessments receive
    no credit and are reported separately in Appendix~\ref{app:e1-results}.}
\end{table*}

Table~\ref{tab:schema-neutral-construction-results} shows that producing
runnable strategies is more common than completing tasks with all required
behavior verified. QuantHarness achieves the highest Task Success at
90.72\%, compared with 74.23--84.54\% for the comparison systems.
Across systems, Strategy Run Success exceeds Task Success. Aider illustrates
this gap: its strategies execute successfully on 94.85\% of tasks, but its
Task Success is 79.38\%. Its average Strategy Requirement Satisfaction reaches
99.09\%, while Strategy Runtime-Evidence Coverage is 92.64\%.
These results show that successful execution and high average requirement
satisfaction can coexist with incomplete runtime verification.
Reliable completion requires the execution, requirement, and evidence
conditions to hold together within each task.

QuantHarness more consistently meets requirements and provides runtime evidence
while encountering fewer failures during construction and execution. 
OpenHands has the lowest Data Pipeline Success and Strategy Run Success,
at 86.60\% and 84.54\%, respectively; Appendix~\ref{app:e1-results} records
13 generation or data-pipeline failures, supporting construction as its
main bottleneck. Codex CLI shows a gap between Strategy Requirement
Satisfaction (97.50\%) and Strategy Runtime-Evidence Coverage (91.11\%),
indicating less complete evidence than its implementation-level assessments
suggest. SWE-agent likewise achieves 93.81\% Strategy Run Success but
84.54\% Task Success; the case-level breakdown identifies five mechanically
completed tasks with unmet or unverifiable requirements and four with
missing runtime evidence. For Aider, the same breakdown records ten
mechanically completed tasks with unmet or unverifiable requirements,
explaining why its high average criterion scores do not imply complete
requirement fulfillment within individual tasks.
These differences are consistent with QuantHarness's Workflow: the strategy
specification makes the request explicit, data grounding checks the inputs
before implementation, and validation and review connect candidate acceptance
to supporting artifacts.

Two QuantHarness cases illustrate the difference between an unmet requirement
and missing runtime evidence. In the industry-aware multifactor task,
selection without industry constraints is permitted when industry labels
are unavailable, provided that this limitation is reported. The code
implements the fallback but omits the required reporting, leaving an explicit
requirement unmet. In the relative-strength task, the strategy implements
a cash fallback when no stock passes the liquidity and trend filters.
Both strategies pass validation and execution, but one omits required
reporting while the other leaves a conditional behavior unverified.
Further details are provided in Appendices~\ref{app:benchmark-details} and~\ref{app:e1-results}.

\subsection{Harness Boundary Ablation and Stress Tests}
\label{subsec:harness-ablations}

The complete-system comparison shows QuantHarness's overall reliability, but
not which parts of Harness Core produce it. We therefore examine how the Harness
selects information for each stage and controls tool use and changes to the run.
For each test, we compare the
system with the target boundary enabled against its ablated configuration,
while keeping the task, shared inputs, and remaining boundary settings fixed.
We report paired results on a fixed subset of 30 FinRelBench tasks, selected
before the experiments to cover diverse strategy rules and both daily-score
and minute-order outputs.

\begin{table*}[t]
  \caption{Paired boundary ablations and stress tests on 30 FinRelBench tasks.}
  \label{tab:harness-boundaries}
  \centering
  \scriptsize
  \setlength{\tabcolsep}{3.5pt}
  \begin{tabular*}{\textwidth}{@{\extracolsep{\fill}}llrr@{}}
    \toprule
    \multicolumn{4}{l}{\textit{A. Normal runs}} \\
    Boundary & Successful outcome $\uparrow$ & On & Off \\
    \midrule
    Information & Initial implementation succeeds & 25/30 & 23/30 \\
    Action & Workflow completes correctly & 30/30 & 23/30 \\
    \addlinespace[3pt]
    \multicolumn{4}{l}{\textit{B. Controlled stress tests}} \\
    Boundary & Observed failure $\downarrow$ & On & Off \\
    \midrule
    Information & Invalid information used & 0/30 & 11/30 \\
    Action & Unapproved file created & 0/30 & 22/30 \\
    Commitment & Changed candidate frozen & 0/30 & 14/30 \\
    \bottomrule
  \end{tabular*}
\end{table*}

In normal runs, the Information Boundary controls the history given to the
Agent; the Action boundary limits its tools and checks calls before execution.
For the Information ablation, both variants draw from the same source run
history and receive the same input-length budget: with the boundary enabled,
the Agent receives the requirements and implementation decisions that still
apply; when it is weakened, the Agent receives a chronological slice and must
decide for itself which information remains valid. With Action enabled, the
Agent sees only the tools needed at that stage and calls are checked; after
ablation, the Agent sees more tools, receives no guidance, and faces fewer
checks. Table~\ref{tab:harness-boundaries} Panel~A shows that, with the Information
Boundary enabled, 25/30 initial implementations execute successfully and
satisfy the public requirements, compared with 23/30 under ablation.
With the Action boundary enabled, all 30 Workflows complete correctly,
compared with 23/30 under ablation, where failures include invalid data
bindings and empty outputs. Together, these results suggest that
stage-appropriate information and coordinated tool guidance and execution
checks help agents preserve task requirements and carry development
through to executable strategies.

Normal runs may not naturally expose the failure conditions that each boundary
is designed to prevent, so we next conduct controlled stress tests. Each test
introduces a targeted opportunity for a boundary violation under matched task
and starting-stage conditions, and counts a violation only when saved evidence
confirms it.
For Information, the ablation adds information from superseded and rejected
decisions to information from the accepted decision, leaving the Agent to
decide what remains valid. Table~\ref{tab:harness-boundaries} Panel~B shows that
no run uses invalid information with the Information Boundary enabled,
compared with 11/30 cases under ablation.
For Action, the Agent may request data construction before approval to avoid a
synthetic scheduling delay. It chooses early execution in 22/30 cases; each
request is then shared between the two variants. The Action Boundary blocks all
22 requests, while removing the check allows all 22 to create files. For
Commitment, we preselect 14 cases without using task content or results, replace
the candidate between Judge approval and freeze, and leave the other 16 as
controls; the ablation removes the identity check. The check catches all 14
replacements; after ablation, all 14 reach freeze,
while the controls remain unchanged. Together, these checks keep invalid
information, unauthorized writes, and changed candidates from carrying forward
into later stages. Appendix~\ref{app:e2-results} describes the boundary-test protocols and
stress-test activation criteria.

\subsection{Auditability and Replay}
\label{subsec:auditability-replay}

\begin{table}[t]
  \caption{Evidence retrieval and replay for 44 run records covering 30 FinRelBench tasks.}
  \label{tab:artifact-audit}
  \centering
  \scriptsize
  \setlength{\tabcolsep}{4pt}
  \begin{tabularx}{\columnwidth}{@{}>{\raggedright\arraybackslash}Xrr@{}}
    \toprule
    Evidence view & Exact evidence & Replay \\
    \midrule
    Organized artifacts & 100.0\% & 44/44 \\
    Same artifacts, unorganized & 43.2\% & 0/44 \\
    Final artifacts only & 27.6\% & 0/44 \\
    \bottomrule
  \end{tabularx}
\end{table}

The preceding tests assess boundary behavior during execution, but they do not
show whether the \emph{Artifact Store} preserves run evidence for later
inspection and decision replay. We use 44 run records produced by the Commitment
stress test, covering the same 30 FinRelBench tasks. We give an LLM three views
of each record, with the same lookup budget for each: artifacts organized by Workflow
stage and their links, the same visible artifacts without that organization,
and final artifacts only. We test whether the auditor returns the exact location
and matching hash for requested evidence. Replay reruns validation and
backtesting in isolation and uses the saved Judge proposal to reconstruct
Commitment authorization outcomes and terminal candidate identity.

Table~\ref{tab:artifact-audit} shows that the organized view recovers 100.0\%
and replays 44/44; the unorganized and final-only views recover 43.2\% and
27.6\%, respectively, and replay none. These results show that saving files
alone is not enough: their organization and intermediate artifacts enable later
inspection and replay. Appendix~\ref{app:e2-results} describes the record views, evidence-retrieval
protocol, and conjunctive decision-replay criterion.

\FloatBarrier

\section{Limitations and Future Work}
\label{sec:limitations}

Using GPT-5.5 as a shared generation backbone helps distinguish system-level differences 
from variation due to model choice. This common setting supports a controlled comparison, 
although the findings remain specific to a single generation model. Broader model coverage would 
require substantial additional resources, as each task involves a complete development 
trajectory with multiple stages of generation and evaluation. Extending the study to other 
model families and capability levels would help establish whether the observed reliability gains 
generalize and clarify the trade-off between reliable completion and inference cost.

FinRelBench currently evaluates explicitly specified rule-based tasks in a
long-only A-share environment through daily-score and minute-order interfaces.
These tasks connect strategy requirements to verifiable data processing and
execution behavior under a shared market protocol. Extending evaluation to
other markets, asset classes, and more open-ended research tasks would test
development reliability under different data conventions and trading
constraints. Additional historical periods would further broaden the
conditions covered by this evaluation.

Our evaluation of Harness Core focuses on governance within individual
development trajectories. Building on the retained run records, a further
direction is governed experience reuse across tasks and runs. A memory module
could enable later runs to reuse verified data bindings, recurring failure
patterns, and successful repair procedures, with Harness Core mediating memory
writes and retrieval. Retrieved experience would remain linked to its source
evidence and be checked for compatibility with the current task and market
context before use. Evaluating this extension would examine whether experience
reuse improves Task Success and efficiency while preserving traceability and
the separation between development and held-out evaluation.

\section{Conclusion}
\label{sec:conclusion}

This paper presents QuantHarness, a two-layer harness for governing the
development trajectory of LLM-driven quantitative strategies. Its
finance-specific Workflow structures strategy construction and evaluation,
while Harness Core mediates information handoff, tool actions, and lifecycle
commitments through the Information, Action, and Commitment boundaries. An
artifact-backed run record links each terminal outcome to the evidence and
decisions that produced it, and FinRelBench evaluates strategy-development
reliability through executability, requirement satisfaction, and runtime
evidence. The experiments show improved reliable completion on FinRelBench,
containment of targeted workflow violations, and stronger evidence recovery
and decision replay from structured provenance. Together, these results expose
the gap between producing a runnable strategy and establishing that the
requested strategy was faithfully developed through an inspectable process.
QuantHarness operationalizes a reliability-first view in which strategy
development is assessed through task fidelity, process control, and supporting
evidence, providing a systems foundation for more rigorous financial-agent
research.

\bibliographystyle{ACM-Reference-Format}
\bibliography{references}

% Preserve a readable single-column presentation for the technical appendices.
% Page numbers continue from the paper; section labels become A--F.
\onecolumn
\appendix
\fontsize{11}{13.2}\selectfont
\raggedbottom
\section*{Appendices}
\pdfbookmark[0]{Appendices}{appendices}
\begin{description}
  \item[\ref{app:benchmark-details}] \hyperref[app:benchmark-details]{FinRelBench Details and Evaluation Protocol}
  \item[\ref{app:e1-results}] \hyperref[app:e1-results]{Additional Reliability Results and Failure Analysis}
  \item[\ref{app:e2-results}] \hyperref[app:e2-results]{Boundary Tests, Auditability, and Replay}
  \item[\ref{app:finresearchsuite}] \hyperref[app:finresearchsuite]{FinResearchSuite}
  \item[\ref{app:e3-results}] \hyperref[app:e3-results]{FinResearchSuite Results}
  \item[\ref{app:implementation}] \hyperref[app:implementation]{Implementation and Reproducibility Details}
\end{description}

\section{FinRelBench Details and Evaluation Protocol}
\label{app:benchmark-details}

FinRelBench evaluates whether a strategy-development system turns an explicit
request into an executable implementation supported by requirement-level
evidence. This appendix describes the task coverage, scoring rules, worked
examples, and external evaluation procedure behind the main comparison.

\subsection{Benchmark Composition and Coverage}
\label{app:benchmark-case-inventory}

FinRelBench contains 97 explicit rule-based tasks: 64 daily-score tasks and 33
intraday-order tasks. Each public case contains an identifier, a
natural-language query, and an output type. The 22 auxiliary research tasks in
Appendix~\ref{app:finresearchsuite} are evaluated separately, not pooled with
this benchmark.

\paragraph{Original tasks.}
The fourteen daily-score cases cover moving-average crossover, filtered
reversal, medium-term risk-adjusted momentum, volatility-regime mean reversion,
multifactor selection, industry-aware multifactor selection, low-volatility
portfolio construction, momentum with volatility and cash gating,
relative-strength rotation, index-gated selection, price-limit-distance
filtering, gap reversal, factor-bucket rotation, and rolling-pattern ranking.
The three intraday-order cases cover opening-range breakout, execution-aware
target orders, and minute-level liquidity-shock reversal. These tasks introduce
explicit requirements concerning adjusted prices, security and timestamp
identity, tradability, regime filters, missing metadata, and cash fallback.

\paragraph{Adapted tasks and scope.}
The remaining 80 tasks contain 50 daily-score and 30 intraday-order requests,
adapted from \emph{QuantCode-Bench}, distributed under the MIT license.
Selection retains explicit rule-based strategies whose core trading logic can
be expressed as verifiable requirements using the available daily or minute
equity data under long-only trading. Tasks whose defining requirements depend
on unavailable inputs, non-equity instruments, short selling, or machine
learning are excluded. Adaptation preserves the core trading logic and retains
numerical parameters where compatible, while aligning the stock universe,
data interfaces, and execution timing with A-share trading rules, including
T+1 settlement.

The adapted tasks broaden indicator combinations, lookbacks, stateful entry
and exit rules, and multi-frequency timing. Examples include 252-session
absolute momentum, volume-and-RSI breakouts, re-entry following an earlier exit,
Donchian contraction breakouts, and session-VWAP reversals. Minute tasks may
construct session-aligned 5-, 15-, or 30-minute aggregates but still emit orders
through the same latency-constrained interface.

\begin{table}[htbp]
  \caption{Composition of the current FinRelBench task set.}
  \label{tab:finrelbench-composition}
  \centering\small
  \begin{tabular}{lrrr}
    \toprule
    Origin & Daily scores & Intraday orders & Total \\
    \midrule
    Original reliability-oriented tasks & 14 & 3 & 17 \\
    Adapted QuantCode-Bench tasks & 50 & 30 & 80 \\
    Total & 64 & 33 & 97 \\
    \bottomrule
  \end{tabular}
\end{table}

Table~\ref{tab:finrelbench-composition} summarizes the tasks by origin and
output type; Appendix~\ref{app:benchmark-worked-examples} gives representative
requests and their evaluation. Public requests state the strategy requirements,
while source mappings and evaluator-specific rubrics are reserved for
evaluation. Results are grouped by task origin and output type to reflect
these two complementary dimensions of coverage.

\subsection{Rubric Construction and Reliability Endpoints}
\label{app:rulebench-rubric}

To assess whether an implementation fulfills a public request, we use a
task-specific \emph{rubric}: a structured evaluation specification that maps
the request and the validity conditions needed for correct implementation to
criteria and disqualifying conditions. For task $i$, we write
$\mathcal R_i=(\mathcal C_i^d,\mathcal C_i^s,\mathcal F_i^d,\mathcal F_i^s)$,
where $d$ and $s$ denote the data and strategy dimensions, $\mathcal C$ contains
criteria, and $\mathcal F$ contains predefined forbidden conditions.
Data criteria concern security and timestamp identity, required inputs,
temporal coverage, and their actual use; strategy criteria concern signals,
portfolios, timing, and execution. Forbidden conditions identify invalid
behavior, such as using information unavailable at decision time.

For each task, we decompose the public request into independently assessable
data and strategy obligations. Compound instructions are separated when their
components can be satisfied or violated individually; for example, transaction
costs and slippage are assessed as separate criteria. Conditional obligations
retain both the trigger and its required response, so decomposition preserves
the meaning of the requested behavior.
Each criterion has an identifier and is traceable either to an explicit public
requirement or to a necessary semantic precondition---a validity condition
required for correct implementation of the task. Each added precondition records
a rationale explaining its necessity, such as sufficient historical observations
to compute a requested moving average.

To reflect each criterion's importance to the requested strategy and the
validity of its inputs, we assign a task-specific \emph{must} or \emph{should}
level. Must criteria capture requirements treated as indispensable to the
task's defining data or trading semantics and must all hold for minimum
semantic acceptance. Should criteria capture supporting requirements that
refine the implementation; partial satisfaction is permitted at this minimum
gate. This distinction sets the minimum acceptance standard without changing
the requirements stated in the public request.
For each dimension $k\in\{d,s\}$, $\mathcal C_i^k$
is partitioned into the disjoint sets of must criteria $\mathcal M_i^k$ and
should criteria $\mathcal H_i^k$. These levels determine the minimum semantic
gate formalized below. Each dimension contains at least one must criterion,
so $|\mathcal C_i^k|>0$. If $\mathcal H_i^k$ is empty, the should-satisfaction
condition is automatically met.

To assess whether judgments remain stable when only wording changes, each
criterion has two semantically equivalent statements. These rephrase the same
atomic requirement without changing what is assessed, when it applies, the
required behavior, or its numerical constraints. For illustration,
``The portfolio holds at most 20 securities'' and ``The number of securities
held in the portfolio does not exceed 20'' express the same constraint:
the criterion identifier, must/should level, and 20-security limit remain
unchanged. To check rubric completeness and source traceability, structural
validation verifies task coverage, identifiers and links, and exact matches
between source quotations and public requests. The two statements share one
criterion identifier and level, while automated checks verify numerical
consistency between them. Source quotations remain verbatim even when the
criterion wording changes. These checks establish structural and numerical
consistency, not semantic equivalence by themselves. The rubric definitions
are frozen before evaluation; Appendix~\ref{app:semantic-panel} describes how
the two wordings are used in the evaluation procedure.

To identify which criteria directly encode requirements explicitly stated in
task $i$'s public request, we define
$\mathcal E_i\subseteq\mathcal C_i^d\cup\mathcal C_i^s$ as the subset of all
such data and strategy criteria. The criteria included in $\mathcal E_i$
are determined during rubric construction, before any system output is assessed.
Each criterion in $\mathcal E_i$ is linked to a verbatim quotation from the
public request. The same criterion retains its identifier in both equivalent
forms of the rubric, allowing its judgments to be aligned across the two
wordings. Criteria added solely as necessary semantic preconditions are
excluded from $\mathcal E_i$, although they remain in $\mathcal C_i^d$ or
$\mathcal C_i^s$ for semantic evaluation. For example, when the public request
explicitly requires adjusted prices, the corresponding input criterion belongs
to $\mathcal E_i$, whereas a general check of security and timestamp identity
introduced only as a validity precondition does not. Membership in
$\mathcal E_i$ identifies a criterion's origin within the full criterion set,
whereas its must/should level
specifies its role in the minimum semantic gate. Membership is fixed during
rubric construction and does not change with the observed run outcome. A
criterion corresponding to an explicit conditional requirement remains in
$\mathcal E_i$ even when the triggering condition is not observed during the
run. For complete fulfillment of the public request, Task Success requires
every criterion in $\mathcal E_i$ to be both satisfied and supported by runtime
evidence, as formalized below.

For every criterion in $\mathcal C_i^d\cup\mathcal C_i^s$, we record two
assessments: semantic status evaluates whether the requirement is met, while
evidence strength distinguishes runtime support for the relevant behavior
from code or declarations alone.
An individual reviewer assigns semantic status as \textsc{Satisfied} when
evidence supports fulfillment, \textsc{Not Satisfied} when evidence shows
contradiction or omission, and \textsc{Not Verifiable} when evidence is
insufficient or ambiguous.
Evidence strength records the implementation support available for the
criterion, using the labels \textsc{Operationalized}, \textsc{Code Only},
\textsc{Declared Only}, and \textsc{Not Verifiable}. Operationalized evidence
requires executable code on the actual launch path together with execution or
artifact evidence of the relevant behavior. Code Only denotes a visible
implementation without a corresponding runtime witness. Declared Only denotes
an unsupported claim in comments or metadata, or a setting unconnected to
execution. Not Verifiable evidence strength denotes the absence of defensible
implementation evidence.

Data and strategy assessments attribute defects to the relevant semantic
dimension, avoiding duplicate semantic penalties for the same missing input.
Insufficient data do not by themselves invalidate
strategy logic correctly expressed in code, although they may limit runtime
evidence. For a strategy criterion that explicitly requires observed end-to-end
behavior, code-level correctness alone is insufficient to establish satisfaction.

The external panel combines these individual judgments using the procedure in
Appendix~\ref{app:semantic-panel}. A semantic-status disagreement that remains
unresolved is recorded as \textsc{Disagreed}, a panel-level outcome rather than
an individual reviewer's judgment. For a fixed evaluated system and its run on
task $i$, let $s_{ic}=1$ when criterion $c$ has final status Satisfied, and
$e_{ic}=1$ when its evidence strength is Operationalized and its final status
is not Disagreed, with each indicator zero otherwise.
The system index is omitted throughout the following run-level definitions.
These indicators capture different properties: an incorrect rule can be
visibly executed, while a correct conditional branch can lack runtime evidence.

Before applying the semantic gate, we check whether each dimension has the
minimum materials needed for assessment. The data dimension is reviewable when
the collected evidence records at least one dataset artifact as present.
The strategy dimension is reviewable when the evidence contains nonempty,
readable source code for the strategy
or dataset construction. Reviewability permits semantic assessment but
does not establish execution success or requirement satisfaction.
For a reviewable assessment, the minimum semantic gate in dimension $k$ is
\begin{equation}
  G_i^k =
  \left(\prod_{c\in\mathcal M_i^k}s_{ic}\right)
  \mathbf{1}\!\left[
    \sum_{c\in\mathcal H_i^k}s_{ic}
    \ge \left\lceil 0.75|\mathcal H_i^k|\right\rceil
  \right](1-v_i^k),
  \label{eq:rubric-minimum-gate}
\end{equation}
where $\mathbf{1}[\cdot]$ equals one when the enclosed condition holds and zero
otherwise. To account for task-invalidating defects beyond the listed items,
an \emph{unlisted fatal defect} is a directly evidenced defect in the current
dimension that is not covered by any listed criterion or forbidden condition
but would invalidate the task implementation even if all listed items passed.
Such a defect requires confirmation through the blind procedure in
Appendix~\ref{app:semantic-panel}. The flag $v_i^k$ is one when a forbidden
condition in $\mathcal F_i^k$ is confirmed or an unlisted fatal defect is
confirmed through that procedure, and zero otherwise. Forbidden conditions
are assessed as present, not present, or not verifiable. Thus the gate requires
all must criteria, at least 75\% of should criteria rounded up, and no confirmed
disqualifying defect. If dimension $k$ is not reviewable, we set $G_i^k=0$.

Passing the minimum semantic gates does not by itself establish complete
fulfillment of the public request. Task Success therefore requires both gates
to pass and both the data pipeline and strategy to execute successfully;
every criterion in $\mathcal E_i$ must also be satisfied and supported by
runtime evidence. \emph{Data Pipeline Success} requires the dataset to be
constructed, saved, and consumed by the strategy; \emph{Strategy Run Success}
requires the strategy to complete execution and produce its declared output.
Let $P_i=1$ when both execution checks pass, and zero otherwise. Then
\begin{equation}
  \mathrm{TS}_i = P_i\,G_i^d\,G_i^s\,
  \mathbf{1}\!\left[s_{ic}=e_{ic}=1\ \text{for all }c\in\mathcal E_i\right].
  \label{eq:rubric-task-success}
\end{equation}
Every explicit requirement must therefore be satisfied and operationalized,
even when its rubric level is should.

\begin{samepage}
Task Success gives a binary overall outcome. To describe requirement
satisfaction and runtime-evidence coverage, including for runs that do not
achieve Task Success, we separately aggregate the previously defined $s_{ic}$
and $e_{ic}$ within each task and dimension. These fractions are
\emph{Requirement Satisfaction} (RS) and \emph{Runtime-Evidence Coverage} (REC),
respectively:
\begin{equation}
  \mathrm{RS}_i^k =
    \frac{\sum_{c\in\mathcal C_i^k}s_{ic}}{|\mathcal C_i^k|},
  \qquad
  \mathrm{REC}_i^k =
    \frac{\sum_{c\in\mathcal C_i^k}e_{ic}}{|\mathcal C_i^k|}.
  \label{eq:rubric-diagnostics}
\end{equation}
\end{samepage}
RS and REC measure the extent of criterion satisfaction and runtime-evidence
coverage, with scoring distinct from the minimum semantic gate and execution
checks. Although must and should criteria play different roles in that gate,
each criterion receives equal weight in these two fractions. Passing an
execution check likewise does not establish runtime evidence for every
requirement: a zero-row output, for example, can pass a launch check while
leaving a conditional requirement without runtime evidence.
Appendix~\ref{app:benchmark-aggregation} specifies how these task-level values
are aggregated across tasks and how system failures and incomplete assessments
are handled.

\subsection{From Public Request to Evidence and Verdict}
\label{app:benchmark-worked-examples}

The following examples apply the same rubric definitions in
Appendix~\ref{app:rulebench-rubric} to two contrasting QuantHarness runs.
In the first, a conditional requirement is semantically satisfied but lacks a
runtime witness; in the second, requirement satisfaction and runtime support
are complete. Tracing each public request through data and strategy evidence
to the final verdict shows why runs that pass both execution checks and
minimum semantic gates can still differ in Task Success.

\subsubsection{Daily Relative-Strength Rotation}

The first run illustrates the distinction between implementing a conditional
rule and providing evidence of its execution. Its public request is:
\begin{quote}\small
Build a long-only relative strength rotation strategy across a small eligible
equity universe. Rank names by recent relative performance, hold the strongest
names, rebalance weekly or monthly, and stay in cash when no name passes
liquidity and trend filters.
\end{quote}

To assess whether the data support the requested ranking and filtering, the
rubric checks security identity, comparable price histories, and liquidity and
trend inputs. The persisted daily dataset contains 500 securities over
90 trading dates, with 44,936 rows and no
duplicate entity--time pairs. Its profiles show aligned price and momentum
inputs, alongside liquidity and trend features consumed by the strategy.
On this evidence, the resolved panel marks all three data criteria, including
the shared security-identity criterion, \textsc{Satisfied}/\textsc{Operationalized}.
Each therefore has $s_{ic}=e_{ic}=1$, giving
$\mathrm{RS}_i^d=\mathrm{REC}_i^d=3/3=100\%$ under
Eq.~\eqref{eq:rubric-diagnostics}.

With these inputs established, the strategy rubric checks ranking (S01),
selection of the strongest names (S02), cash fallback (S03), and weekly or
monthly rebalancing (S04). All four come directly from the public request and
therefore belong to $\mathcal E_i$.
For S01, S02, and S04, the executed strategy combines recent momentum measures
into relative-strength scores, selects the highest-ranked eligible names,
and schedules signals on the last trading day of each week. The reproduced
output contains positive scores for the selected names, with sampled signal
dates one week apart. The panel therefore marks these three criteria
\textsc{Satisfied}/\textsc{Operationalized}, with $s_{ic}=e_{ic}=1$.

S03 warrants closer examination because its implementation is supported but
its conditional behavior lacks a runtime witness. In the executable code,
eligibility requires both liquidity and trend filters; if no name passes,
the branch emits no scores for that date. The effective execution configuration
disables retention of prior holdings when scores are absent. These code and
configuration facts support the cash-fallback rule, but do not establish that
the condition was triggered. The supplied run evidence does not identify a
date on which no name passed both filters. The resolved panel therefore
records S03 as
\textsc{Satisfied}/\textsc{Code Only}: $s_{i,\mathrm{S03}}=1$ but
$e_{i,\mathrm{S03}}=0$. Thus all four strategy criteria are satisfied, while
three have runtime support, giving $\mathrm{RS}_i^s=4/4=100\%$ and
$\mathrm{REC}_i^s=3/4=75\%$.

For the final task verdict, the execution checks complement these
criterion-level judgments: the datasets were constructed, saved, and consumed,
and the strategy completed with its declared score output, so $P_i=1$.
Both semantic dimensions are reviewable, all must and should
criteria are satisfied, and the panel records the predefined forbidden
conditions as not present with no confirmed unlisted fatal defect. Hence
both minimum semantic gates pass: $G_i^d=G_i^s=1$.
Despite these passes, S03 belongs to $\mathcal E_i$ and lacks runtime evidence,
so Eq.~\eqref{eq:rubric-task-success} yields $\mathrm{TS}_i=0$. The missing
runtime witness, rather than an incorrect cash rule, separates minimum semantic
acceptance from complete task success in this example.

\subsubsection{Daily Adjusted-Price Moving-Average Crossover}

The second run provides the complementary case: the requested behavior is
both implemented and supported by runtime evidence. Its public request is:
\begin{quote}\small
Implement a daily moving-average crossover strategy using adjusted close
prices. Compute 20-day and 60-day moving averages. Generate positive scores
when the short average is above the long average, rebalance on the next
trading session, hold at most 20 names, equal weight positions, and include
transaction costs and slippage.
\end{quote}

Here, the data assessment checks whether each security has adjusted-price
inputs and sufficient chronological history to compute the requested moving
averages. The persisted dataset preserves individual
securities and dated observations, with 80--90 observations per security.
The implementation derives adjusted close from the bound close-price and
adjustment-factor columns. Full-window rolling calculations require sufficient
observations before producing a signal.
The dataset profiles and reproduced output support the use of these inputs
and the required history, so the resolved panel marks all three data criteria
\textsc{Satisfied}/\textsc{Operationalized}. Each has $s_{ic}=e_{ic}=1$,
giving $\mathrm{RS}_i^d=\mathrm{REC}_i^d=3/3=100\%$ under
Eq.~\eqref{eq:rubric-diagnostics}.

Using these inputs, the executed strategy computes the requested 20-day and
60-day adjusted-close
averages (S01) and emits positive scores only when the short average exceeds
the long average (S02); the reproduced output records the resulting selections.
The code caps selections at 20 names per signal date, and the reproduced
output respects this limit (S04).

For the resulting selections, the effective execution configuration specifies
a one-session lag (S03) and equal-weight portfolio construction (S05).
Dated orders and fills support next-session execution, while the orders'
equal-weight targets support the weighting rule. Fee deductions and
slippage-adjusted fill prices provide runtime evidence for transaction costs
(S06) and slippage (S07). All seven strategy criteria directly encode public
requirements and belong to $\mathcal E_i$.
On this combined implementation and execution evidence, the panel marks all
seven strategy criteria \textsc{Satisfied}/\textsc{Operationalized}, with
$s_{ic}=e_{ic}=1$ for each. Thus
$\mathrm{RS}_i^s=\mathrm{REC}_i^s=7/7=100\%$.

For the final task verdict, the execution checks again complement these
criterion-level judgments: dataset construction, persistence, and consumption
succeed, followed by strategy completion with its declared score output,
giving $P_i=1$. Both semantic dimensions are reviewable, all must and
should criteria are satisfied, and the panel records the predefined forbidden
conditions as not present with no confirmed unlisted fatal defect. Therefore
$G_i^d=G_i^s=1$. Unlike the first run, every explicit requirement also has
both a satisfied status and operationalized evidence. With all conditions
met, Eq.~\eqref{eq:rubric-task-success} yields $\mathrm{TS}_i=1$.

Together, the examples distinguish successful execution, semantic satisfaction,
and complete runtime evidence: the first two do not by themselves establish
the third. Both runs pass the execution checks and minimum semantic gates;
their different Task Success verdicts reflect whether every explicit
requirement has a runtime witness under the same evaluation rule. The next
subsection describes the external panel that produces these criterion-level
judgments and how it resolves disagreements.

\subsection{External Panel Resolution and Calibration}
\label{app:semantic-panel}

The criterion-level judgments illustrated above are produced by an external
benchmark panel, separate from the internal Workflow Judge. The internal Judge
recommends acceptance or recovery during development; the external panel
evaluates submitted evidence without feeding its judgments back into further
candidate development.

The panel uses the two equivalent rubric wordings described in
Appendix~\ref{app:rulebench-rubric} and frozen before evaluation to assess
the stability of judgments across equivalent formulations. Under each wording,
Claude Opus 4.6 (\texttt{claude-opus-4-6}) and Gemini 3.1 Pro
(\texttt{gemini-3.1-pro-preview}) independently assess the same anonymized
evidence. They return criterion-level semantic status, evidence strength, and
evidence locators rather than an overall pass or score.

When these independent judgments disagree on decisions relevant to minimum
semantic acceptance or complete fulfillment of the public request,
DeepSeek V4 Pro (\texttt{deepseek-v4-pro}) provides
targeted arbitration. Within a wording, the overlapping triggers concern
must-criterion satisfaction; whether the number of satisfied should criteria
reaches the fixed threshold while minimum acceptance remains possible;
the semantic status or evidence strength of a criterion in the
explicit-requirement set $\mathcal E_i$;
or whether a predefined forbidden condition is present. For criteria not sent
to arbitration, an agreed semantic status is retained; otherwise, the status is
\textsc{Disagreed}. In either case, the weaker of the two evidence-strength
labels is retained.

After this within-wording aggregation, the same arbiter addresses any remaining
differences between the two sets of results. When a criterion has the same
semantic status and evidence strength under both wordings, those labels are
retained, including a shared Disagreed status. In both arbitration stages, the
arbiter judges the original evidence without seeing previous conclusions.
An unlisted fatal defect, as defined in Appendix~\ref{app:rulebench-rubric},
additionally requires blind confirmation as an independent defect in the
relevant dimension, not already covered by a listed criterion or forbidden
condition. The final labels and execution checks then determine the
reliability endpoints using the fixed formulas in that appendix; any criterion
with a final Disagreed status receives zero credit in both RS and REC.

To check whether this review procedure identifies fulfilled requirements and
defects as intended, we use controlled cases with known expected judgments.
Calibration here means verification against these expected labels. The 17 anonymized
\emph{static fixtures} are preconstructed, fixed evidence bundles containing
public requests, code excerpts, dataset and output profiles, and execution
records, with expected judgments specified separately. Four positive controls
provide implementations expected to satisfy the requirements, covering daily
crossover, opening-range breakout, adapted daily momentum, and adapted intraday
breakout. Thirteen negative controls introduce defects in three groups:
data adequacy and identity (insufficient history, incorrect observation
identity, collapsed entities, unusable timestamps, unadjusted prices, and
incorrect intraday frequency); strategy logic and timing (reversed rules,
entry during range formation, an off-by-one lag, and omission of the required
pre-entry channel-contraction condition); and execution constraints
(same-session daily execution, T+1 violation, and zero execution latency).

For example, the insufficient-history fixture supplies 30 observations for a
60-session requirement. The long-window average remains undefined, while the
supplied execution record reports normal termination with an empty output.
This tests whether the panel attributes the defect to inadequate data coverage.
Across the full fixture set, the final-panel checks require agreement with the
expected must statuses, specified evidence strengths for must criteria,
forbidden-condition and unlisted-fatal decisions, and, where specified, whether
all explicit requirements are satisfied and operationalized. The saved reports
record successful completion of these predefined core checks across the 17
fixtures, with should-label comparisons reported separately.

To make these checks traceable to the evaluated configuration, each validation
report records hashes of the protocol, rubric, judge configuration, output
schema, instructions, and fixture bundle. Appendix~\ref{app:e1-results}
complements these controlled checks with panel consistency and arbitration
statistics on benchmark submissions. The next subsection describes how evidence
from the evaluated systems' benchmark runs is collected and organized for this
review.

\subsection{Evidence Collection}
\label{app:benchmark-evidence}

All evaluated systems follow the same neutral submission rules: declared
data-construction and strategy-execution commands, persisted datasets and
outputs, code on the actual launch path, and captured execution evidence.
These requirements do not oblige other systems to emit QuantHarness stage
names or internal contracts.

The evaluator organizes these materials into an anonymized, structured
evidence pack, supplied to the panel together with the task requirements and
rubric. Dataset profiles, observed file access, executed outputs, and code
references carry evidence labels and locators, allowing criterion-level
judgments to point to specific code or records. Where orders, fills, fees, and
holdings are exported from an existing native run, the export protocol uses
code and artifact hashes, invocation settings, and backtest metadata to
associate the records with their originating execution. This association
preserves their status as evidence from the original execution: exporting or
linking the records does not constitute an independent replay or by itself
establish that a strategy requirement is satisfied. Self-reported fields
provide auxiliary context rather than substitutes for actual execution
materials. The same distinction between reported claims and observed evidence
applies to every system.

An interruption of the evidence-collection infrastructure leaves the assessment
incomplete, rather than confirming a strategy failure.
Appendix~\ref{app:benchmark-aggregation} specifies how
completed and incomplete task assessments enter the aggregate metrics.

\subsection{Aggregation and Missing Outcomes}
\label{app:benchmark-aggregation}

System-level results aggregate the task-level endpoints in
Appendix~\ref{app:rulebench-rubric} across all 97 planned cases for every system.
Data Pipeline Success, Strategy Run Success, and Task Success are each the
number of tasks passing the corresponding check divided by 97.
Requirement Satisfaction (RS) and Runtime-Evidence Coverage (REC) use the
task-level fractions defined in Eq.~\ref{eq:rubric-diagnostics}, separately
for the data and strategy dimensions. Each system-level value is the sum of
the corresponding task values divided by 97, with unavailable values treated
as specified below. This is an unweighted task-macro average: cases with longer
rubrics do not receive greater weight. Must and should criteria have equal
diagnostic weight within a case, without changing their roles in the minimum
semantic gate. For illustration, within one dimension, complete coverage on
a four-criterion case and coverage of four of eight criteria on another case
yield a two-case mean of 75\%, not the pooled-criterion value of 8/12.

Eligibility for external semantic assessment requires successful generation,
a valid submission, and an execution-check record without infrastructure
errors; it does not require either mechanical check to pass.
This run-level eligibility check is distinct from the dimension-level
reviewability conditions in Appendix~\ref{app:rulebench-rubric}, which concern
the minimum evidence available for assessment.
Completed, eligible semantic assessments retain their measured RS and REC
values even when Task Success is zero. A completed system failure that
prevents semantic assessment, such as generation failure or an invalid
submission, contributes zero to these diagnostics rather than removing the
task from their denominators.

Eligibility does not guarantee completion: an external assessment is incomplete
when evidence collection or required panel reviews remain unfinished. Such
tasks receive no Task Success credit and contribute zero to RS and REC, while
their observed Data Pipeline Success and Strategy Run Success outcomes are
retained. They remain in all
planned denominators and are reported separately from confirmed strategy
failures, so missing evaluation is not interpreted as evidence of an incorrect
strategy. Appendix~\ref{app:e1-results} reports the corresponding counts and
task-group results.

\section{Additional Reliability Results and Failure Analysis}
\label{app:e1-results}

This appendix expands the main reliability comparison by task group and
failure stage, then connects the aggregate results to individual requirements
and the evidence used to assess them.

\subsection{Task Groups and Evaluation Coverage}

Table~\ref{tab:finrelbench-groups} expands the main reliability comparison of
five systems on 97 planned tasks per system by output type and task origin.
These are two separate partitions of the same task set. Following
Appendix~\ref{app:benchmark-aggregation}, each cell reports confirmed Task
Success over all planned tasks in its group, including unsuccessful tasks
and incomplete assessments in the denominator.

\begin{table}[htbp]
  \caption{Task Success by output type and task origin. Each pair of columns
  partitions the same 97 planned tasks.}
  \label{tab:finrelbench-groups}
  \centering\small
  \begin{tabular}{lrrrr}
    \toprule
    System & \multicolumn{2}{c}{Output type} & \multicolumn{2}{c}{Task origin} \\
    \cmidrule(lr){2-3}\cmidrule(lr){4-5}
    & Daily scores & Intraday orders & Original & Adapted \\
    \midrule
    QuantHarness & 60/64 & 28/33 & 14/17 & 74/80 \\
Codex CLI & 54/64 & 25/33 & 12/17 & 67/80 \\
OpenHands & 49/64 & 23/33 & 12/17 & 60/80 \\
SWE-agent & 56/64 & 26/33 & 14/17 & 68/80 \\
Aider & 52/64 & 25/33 & 14/17 & 63/80 \\
\bottomrule

  \end{tabular}
\end{table}

QuantHarness has the highest confirmed Task Success in both output groups,
60/64 daily-score tasks and 28/33 intraday-order tasks. The other systems
achieve 49--56 daily-score successes and 23--26 intraday-order successes. In
the original subset, QuantHarness, SWE-agent, and Aider each reach 14/17. In
the adapted subset, QuantHarness reaches 74/80, compared with 60--68 successes
for the other systems. The adapted tasks thus further distinguish systems
that are tied on the original subset.

\paragraph{Incomplete assessments.}
External assessments remain incomplete for one QuantHarness task because
the external reviewer's evidence references failed validation, and for two
OpenHands tasks because a required review or arbitration stage timed out.
In each case, the affected evaluation stage remained unsuccessful after
three attempts. No incomplete external assessments are recorded for Codex CLI,
SWE-agent, or Aider. The three tasks with incomplete assessments remain in the
planned denominators and are reported separately
from confirmed strategy failures in Table~\ref{tab:finrelbench-failures}.

\paragraph{Evaluation configuration.}
All systems are assessed on the same public task set using common rubric and
endpoint definitions. The comparison includes 485 system--task evaluations,
with one retained development trajectory per system--task pair.
Appendix~\ref{app:systems-environments}
specifies the system settings and the budgets for generation and independent
evaluation.

\subsection{Where End-to-End Reliability Breaks Down}

Table~\ref{tab:finrelbench-failures} decomposes the non-successes in the
comparison above into mutually exclusive categories. Incomplete assessments
are separated first; each remaining non-success is assigned to the first
applicable category in the following order: generation or data-pipeline
failure, strategy-execution failure, unmet or unverifiable requirements,
and missing runtime evidence. Each task is counted once, even if it fails
multiple success conditions. By contrast, the separately reported Data
Pipeline Success and Strategy Run Success metrics can both record a failure
for the same task, so their failure counts can overlap.

For the remaining non-successful tasks that pass both the data-pipeline and
strategy-execution checks, the last two categories distinguish requirement
satisfaction from runtime evidence. A task falls
into the unmet or unverifiable requirements category if either minimum
semantic gate fails or any explicit public requirement is not judged
satisfied. In contrast, runtime-evidence-only cases pass both minimum
semantic gates, with every explicit public requirement judged satisfied;
at least one such requirement lacks runtime evidence under the Task Success
definition in Appendix~\ref{app:rulebench-rubric}.

\begin{table}[htbp]
  \caption{Non-success accounting over 97 planned tasks per system. The final
  column is evaluator incompleteness, not a strategy-failure category.}
  \label{tab:finrelbench-failures}
  \centering\small
  \setlength{\tabcolsep}{5pt}
  \begin{tabular}{lrrrrr}
    \toprule
    & \shortstack{Generation/\\data pipeline}
    & \shortstack{Strategy\\execution}
    & \shortstack{Unmet/unverifiable\\requirements}
    & \shortstack{Runtime\\evidence only}
    & \shortstack{Evaluation\\incomplete} \\
    \midrule
    QuantHarness & 0 & 1 & 1 & 6 & 1 \\
Codex CLI & 3 & 5 & 5 & 5 & 0 \\
OpenHands & 13 & 1 & 4 & 5 & 2 \\
SWE-agent & 5 & 1 & 5 & 4 & 0 \\
Aider & 1 & 4 & 10 & 5 & 0 \\
\bottomrule

  \end{tabular}
\end{table}

In the retained results, OpenHands has 13 generation or data-pipeline failures,
whereas QuantHarness has none. For QuantHarness, six of the eight confirmed
failures are evidence-only outcomes, alongside one execution failure and one
unmet reporting requirement. Aider's high average Strategy Requirement
Satisfaction in Table~\ref{tab:schema-neutral-construction-results} likewise
coexists with ten mechanically completed cases with unmet or unverifiable
requirements. These patterns distinguish executable delivery from verified
requirement fulfillment: Task Success requires the execution,
requirement-satisfaction, and runtime-evidence conditions to hold together
within each task. The next subsection illustrates these distinctions through
individual requests, implementations, and supporting evidence.

\subsection{Representative Cases}

The examples below connect the outcome categories above to public requirements,
implementations, and supporting evidence. They illustrate complete requirement
fulfillment, an executed rule defect, and a missing reporting path. The
cash-fallback rule in the relative-strength rotation example in
Appendix~\ref{app:benchmark-worked-examples} provides a complementary illustration
of how a correctly implemented conditional rule can remain without runtime evidence.

\medskip
\noindent\textbf{Case 1. Complete success: adjusted-price moving-average crossover.}
\par\nopagebreak\smallskip
\noindent\textit{System:} QuantHarness \quad
\textit{Task:} \nolinkurl{generic_explicit_ma_crossover}
\par\nopagebreak\smallskip
\noindent
The request specifies 20- and 60-session moving averages of adjusted close
prices, positive scores when the short average is above the long average,
next-session rebalancing, a maximum of 20 equally weighted names, and transaction
costs and slippage. QuantHarness implements these requirements, with evidence connecting
the actual daily-price and adjustment-factor columns to their computation in
the submitted strategy; adjusted close need not already be stored as a separate
input column if it is reproducibly constructed. The final panel marks all
three data and seven strategy criteria
\textsc{Satisfied}/\textsc{Operationalized}, and both mechanical gates pass,
yielding Task Success.

\medskip
\noindent\textbf{Case 2. Runnable but incorrect: ATR pullback in an EMA trend.}
\par\nopagebreak\smallskip
\noindent\textit{System:} Codex CLI \quad
\textit{Task:} \nolinkurl{qcb_daily_atr_pullback_in_ema_trend}
\par\nopagebreak\smallskip
\noindent
The request requires a non-positive score to represent an inactive or exit
state. The executed implementation maintains an \texttt{active} flag that is
set on entry and cleared on exit. However, it additionally requires a positive
ATR-normalized pullback before emitting a positive score, allowing active
states to receive zero scores. The
full output profile contains 54,161 active rows but only 54,021 positive scores,
leaving 140 active rows with zero scores.
The cross-wording arbiter uses this additional condition in the implementation
and the full output counts to mark S06
\textsc{Not Satisfied}/\textsc{Operationalized}. The execution evidence thus
establishes the rule violation, and the task fails despite passing both
mechanical checks.

\medskip
\noindent\textbf{Case 3. Missing required fallback reporting.}
\par\nopagebreak\smallskip
\noindent\textit{System:} QuantHarness \quad
\textit{Task:} \nolinkurl{generic_industry_aware_multifactor}
\par\nopagebreak\smallskip
\noindent
The public request permits a non-industry-constrained fallback only with a record of
unavailable industry metadata. The fallback function returns the selected names
and an industry-metadata availability flag. The caller receives the flag but
does not propagate it to persisted outputs or a limitation report. The
available run contains industry metadata; the unmet requirement concerns the
missing reporting path for the permitted fallback. Although both mechanical
checks pass, the final S03 verdict is
\textsc{Not Satisfied}/\textsc{Code Only}, preventing Task Success.

\subsection{External-Panel Agreement and Arbitration}

The preceding cases illustrate how criterion-level judgments determine task
outcomes. We next examine whether the initial reviewers agree on these
judgments and how often the protocol requires further arbitration.
Table~\ref{tab:finrelbench-panel} summarizes these aspects of the external-panel
protocol in Appendix~\ref{app:semantic-panel} for cases eligible for semantic
assessment with all required reviews complete. Some mechanically unsuccessful
submissions still contain inspectable artifacts and can receive a panel.
The complete-panel count describes review coverage, not Task Success, and does
not alter the all-planned-task denominators in
Appendix~\ref{app:benchmark-aggregation}.

\emph{Exact agreement} is the fraction of criteria for which the two initial
reviewers agree on both semantic status and evidence strength before
arbitration. It uses the canonical wording, the reference version of the two
equivalent rubric wordings. Counts are pooled across must and should criteria
in both data and strategy dimensions and all eligible, complete cases, rather
than averaged per task. The arbitration columns instead report proportions of
these cases: \emph{Any arbitration} includes arbitration within either wording
or across wordings, and \emph{Cross-wording} reports the latter subset. Each case
is counted at most once per arbitration column.

\begin{table}[htbp]
  \caption{External-panel consistency on eligible cases with complete
  assessments. Agreement is criterion-level; arbitration counts cases,
  not model calls. Cross-wording arbitration is a subset of any arbitration.}
  \label{tab:finrelbench-panel}
  \centering\small
  \begin{tabular}{lrrrr}
    \toprule
    System & Complete panels & Exact agreement & Any arbitration & Cross-wording \\
    \midrule
    QuantHarness & 96 & 98.2\% & 25/96 & 8/96 \\
Codex CLI & 95 & 97.5\% & 30/95 & 12/95 \\
OpenHands & 83 & 98.3\% & 20/83 & 9/83 \\
SWE-agent & 92 & 99.4\% & 17/92 & 7/92 \\
Aider & 97 & 98.1\% & 33/97 & 15/97 \\
\bottomrule

  \end{tabular}
\end{table}

Exact agreement ranges from 97.5\% to 99.4\% across systems. Thus, for the large
majority of criteria under the canonical wording, the reviewers independently
reach the same judgment about requirement satisfaction and the strength of
the supporting evidence before arbitration.

This broad agreement does not mean that every task can be assessed without
arbitration. For QuantHarness, 25 of 96 cases invoke arbitration, including
eight with cross-wording arbitration. Each task contains multiple criteria,
so a small fraction of criterion-level disagreements can be distributed across
multiple tasks; arbitration also considers the other wording and differences
between wordings. The table therefore documents both broad initial agreement
under the canonical wording and the task-level use of arbitration in producing
the reported verdicts. These diagnostics characterize the evaluation process,
not strategy quality or independently established judgment accuracy.

\section{Boundary Tests, Auditability, and Replay}
\label{app:e2-results}

The experiments separate three questions: how runtime boundaries affect
ordinary development, whether they contain specified invalid events, and
whether the resulting records support evidence retrieval and decision replay.

\subsection{Case Selection and Shared Experimental Conditions}

Normal-run ablations and controlled stress tests use the same fixed 30-case
subset of FinRelBench: 20 daily-score and ten intraday-order tasks. Cases were
selected from public task text before these experiments to cover different
strategy rules and implementation requirements. The subset contains eight
basic, fourteen intermediate, and eight advanced tasks by assigned
implementation complexity, reflecting differences in rule composition, state
maintenance, and temporal dependencies. Coverage spans moving averages,
momentum, regime filters, multifactor rules, breakouts, and intraday state and
timing constraints, allowing the boundary tests to examine both simple rules
and strategies with interacting conditions.

Within each boundary experiment, Full retains the target boundary and Off
applies its protocol-defined ablation. Paired conditions share the task,
the saved workflow state used as the source checkpoint for that experiment,
the canonical data binding, the market configuration, and the non-target
boundary settings. Saved source artifacts and intervention checks verify that
the paired conditions share these inputs.

The backtest gate uses the 100-stock universe defined by the CSI~100 constituent
snapshot dated 2025-12-31, with a backtest period of 2026-01-05--2026-05-29.
Strategy-generation requests use GPT-5.5 at temperature 0.0. The following
sections specify the stage-specific interventions, endpoints, and units of
analysis for each experiment.

\subsection{Normal-Run Boundary Ablations}
\label{app:normal-boundary-tests}

On the paired tasks described above, normal-run ablations examine how
Information and Action boundaries support development without injected
violations.

\paragraph{Information.}
This ablation tests how organizing prior information into a development
handoff affects code generation.
The shared checkpoint follows task understanding, planning, and the
data-binding agent's proposal and tool evidence. Full receives the resolved
field contract, which specifies the available input fields and their mapping
from data sources to code, together with the implementation plan and compact
development context. Off instead presents earlier workflow records available
to the agent in chronological order, including the task contract,
specification, planning blueprint, raw binding proposal, tool evidence,
Runtime Card, and public execution interface, without the organized context
provided to Full. Both conditions draw on the same earlier records.

For each case, the Full and Off code-generation requests share the same upper
limit on the total number of characters in their instruction text and message
bodies. This shared limit is determined from the Full reference request,
reconstructed from the saved artifacts of the frozen Full run. To stay within
this limit, Off uses a fixed rule that retains nearly equal portions from the
beginning and end of the chronological history. The comparison thus varies
information selection and organization under a common input-length limit.

The intervention changes the context presented to the code-generation model,
while keeping the task requirements and downstream evaluation setup unchanged.
Specifically, the downstream canonical binding, field contract, implementation
plan, quantitative input requirements, and protocol decision remain the same.
All 30 pairs pass the recorded task-contract, data-binding, and
manipulation-validity checks.

We assess the initial candidate separately from the terminal Workflow outcome.
Initial implementation success requires both mechanical execution and
public-obligation conformity for the same task. Mechanical execution
requires the first validation and backtest checks to pass without strategy
repair; the backtest must confirm the expected output protocol and actionable
strategy output, meaning signals or orders usable for trading within the
backtest window. A GPT-5.5 assessment, blinded to the Full/Off condition,
judges the initial source against the predefined public requirements and
recorded execution semantics; public-obligation conformity requires all
assessed requirements to be satisfied. Repaired source does not replace the
initial candidate in this assessment. Terminal completion separately requires
validation, backtest, Judge approval, and authorized freezing with consistent
candidate identity.

\paragraph{Action.}
Full exposes stage-appropriate tools, provides their associated guidance, and
checks authorization at invocation. Off exposes a larger set of registered
tools, removes the corresponding guidance, and weakens the targeted checks.
The ablation changes stage-scoped tool exposure, guidance, and invocation
checks together.

Admissible Workflow completion requires the terminal completion checks above
together with predefined checks on saved planning, data-binding, and
implementation records. These checks verify the expected output protocol,
require predefined minimum numbers of signal-rule, position-rule, risk-rule,
and data-selection entries, and check for required task concepts using
predefined term matches in these records. For example, the moving-average
crossover task requires at least two signal-rule entries and one entry each
for position rules, risk rules, and data selections, together with terms
covering the moving averages, their 20- and 60-day windows, next-session
execution, and equal weighting. These checks assess record structure and
task-concept coverage, rather than the full requirement-level semantics
assessed by FinRelBench.

We separately track unauthorized behavior, since a run can fail to complete
without violating an authorization rule. Observed unauthorized behavior
counts a run if records show a completed unauthorized tool call, an overridden
lifecycle authorization check, or a freeze for which the required checks have
not passed or the frozen artifacts cannot be verified against their recorded
hashes.

\begin{table}[htbp]
  \caption{Normal-run boundary diagnostics on 30 paired tasks, separating
  first-candidate execution and semantics from terminal completion.}
  \label{tab:boundary-normal-details}
  \centering\small
  \begin{tabular}{llrr}
    \toprule
    Boundary & Endpoint & Full & Off \\
    \midrule
    Information & Initial mechanical execution & 29/30 & 30/30 \\
                & Public-obligation conformity & 26/30 & 23/30 \\
                & Initial implementation success & 25/30 & 23/30 \\
                & Terminal completion & 30/30 & 30/30 \\
    Action & Admissible Workflow completion & 30/30 & 23/30 \\
           & Observed unauthorized behavior & 0/30 & 0/30 \\
    \bottomrule
  \end{tabular}
\end{table}

For Information, Table~\ref{tab:boundary-normal-details} reports initial
implementation success on 25/30 tasks under Full and 23/30 under Off, while both
conditions reach 30/30 terminal completion. The initial requirement-conformity
difference remains visible even when both conditions complete the Workflow;
terminal completion does not reassess all public obligations on the final
candidate. For Action, admissible Workflow completion is 30/30 under Full and
23/30 under Off. Neither condition exhibits unauthorized behavior in these
ordinary runs, which need not activate a boundary's rejection condition. The
stress tests below supply explicit opportunities to exercise rejection.

\subsection{Controlled Boundary Stress Tests}
\label{app:boundary-stress-tests}

Table~\ref{tab:harness-boundaries} Panel~B in the main text reports the
aggregate stress-test results. Table~\ref{tab:boundary-stress-protocol}
summarizes the corresponding interventions and evidence criteria.
Information and Action leave the agent
to choose its response, without instructions to take the violating option.
Commitment instead injects an external candidate replacement after approval
to test authorization. All three count an escape only when observable evidence
confirms that the targeted boundary violation occurred.
These tests assess containment of the specified violations, not general
adversarial security.

\begin{table}[htbp]
  \caption{Stress protocols. Counts distinguish exposure, agent activation,
  and verified escape.}
  \label{tab:boundary-stress-protocol}
  \small
  \begin{tabularx}{\textwidth}{@{}p{0.14\textwidth}XX@{}}
    \toprule
    Boundary & Starting condition and intervention & Evidence and denominator \\
    \midrule
    Information &
    A revision handoff contains accepted, superseded, and rejected decisions.
    Full receives the currently accepted version; Off receives the raw versions
    and governance events without an explicit current-version answer. &
    Execution evidence must show use of an obsolete or rejected value.
    Invalid information is used in 0/30 Full and 11/30 Off cases. \\
    \addlinespace
    Action &
    A provisional binding passes mechanical checks but substantive review and
    approval are pending. Building early uses a temporary opportunity for
    parallel work; waiting costs one synthetic scheduling cycle and remains
    feasible. &
    The agent chooses early construction in 22/30 cases. The same request is
    replayed against both guards: 0/22 Full and 22/22 Off requests produce
    unauthorized persistent files. \\
    \addlinespace
    Commitment &
    After a real Judge approval, the candidate implementation is replaced in
    14 prespecified cases; 16 cases retain the approved candidate. Off removes
    the targeted identity enforcement. &
    Compare the hashes of the approved source, the source presented for freezing,
    and the frozen source. Full blocks freezing of all 14 modified candidates,
    whereas Off freezes them; all 16 unchanged controls commit in both
    conditions. \\
    \bottomrule
  \end{tabularx}
\end{table}

\paragraph{Information authority.}
This test examines whether code generation uses the currently accepted
information when the revision history also contains superseded and rejected
versions. Each version links a dataset and its required fields to a derived
feature and execution step. Full receives the currently accepted combination.
Off retains the correct version alongside superseded and rejected versions
and their acceptance, validation, and rollback events, but not the resolved
current-version answer. Public task and data requirements remain unchanged.
The invalid-information endpoint requires execution evidence of using an
obsolete or rejected value, such as recorded access to its fields. The eleven
escapes count verified adoption, separately from other first-pass failures.
This contrast evaluates selection of authoritative information at the revision
handoff.

\paragraph{Action activation and side effects.}
The agent can request dataset construction before the provisional binding has
received substantive review and approval. Building immediately uses a temporary
opportunity to prepare data in parallel with the pending review; waiting for
approval adds one synthetic scheduling cycle but still permits task completion.
The binding may be revised, invalidating early work; the scenario
supplies neither a revision probability nor a recommended choice. The agent
has at most four turns, with a 600-second timeout per model request. It
requests early construction in 22 cases and waits in the remaining eight,
which do not activate the guard. Each early-construction request and its
arguments are shared across Full and Off, using the same real
dataset-construction tool in separate run directories. Invocation-time enforcement blocks the
unauthorized calls in Full; Off allows them to execute, and the resulting files
confirm that unauthorized construction occurred. The all-task escape rate is
0/30 versus 22/30; conditional on the 22 activated requests, it is 0/22 versus
22/22. Two separate
authorized-call controls both pass Full enforcement; these mechanism checks
are reported separately from the 30 agent choices.

\paragraph{Commitment identity.}
This test concerns the transition from Judge approval to freezing: the candidate
may be modified after review, leaving the recorded approval associated with an
earlier version. We test whether the Commitment boundary prevents the modified
candidate from being frozen under that earlier approval. Full and Off share
the recorded Judge approval and a commitment seal that
records the paths and content hashes of the approved candidate and its
supporting artifacts. A fixed hash-based assignment using case identifiers and
a prespecified seed selects the 14 replacement cases, without task-content or
outcome-based selection; the other 16 remain
unchanged. In each exposed pair, the same fixed comment is appended to the
strategy source after approval and sealing, without asking the Judge to
approve a new candidate. This deterministic intervention requires no new
strategy-generation calls. The change preserves executable behavior but alters
byte identity, isolating authorization of the exact reviewed artifact.
An escape requires the frozen source to match the modified live source and
differ from the approved identity recorded in the seal.
Unsupported freezing is 0/30 under Full and 14/30 under Off, or 0/14 versus
14/14 among the modified cases. Both conditions commit all 16 unchanged
controls. The next subsection defines an audit corpus from these clean,
contained, and escaped outcomes.

\subsection{Audit Corpus and Evidence Views}
\label{app:audit-views}

The Commitment stress test above asks whether the system prevents a modified
candidate from being frozen under an earlier approval. We next examine whether
the saved records allow an auditor to recover the evidence needed to understand
what happened during development and how the terminal authorization outcome
arose.

Each audit record links a task's saved development history to one post-Judge
branch of that stress test. The corpus contains 44 records from the same
30 tasks: sixteen in which the candidate remains unchanged after approval and
is frozen (clean); fourteen in which a post-approval replacement is prevented
from being frozen (contained); and fourteen in which the replacement is frozen
under the earlier approval (escaped). Following the prespecified replacement
assignment, the unmodified tasks contribute their Full branches, while the
replacement tasks contribute both their Full (contained) and Off (escaped)
branches. Selection does not depend on audit or replay outcomes.

The audit comparison holds each recorded process and outcome fixed and varies
the evidence available to the auditor. Each record is presented through three
evidence views---organized, flat, and final-only---which specify the artifacts
and organizational information available, giving 132 record--view audits.

For each record, organized and flat views retain the same visible artifact
contents, hashes, and ordering under a shared 240,000-character artifact-content
budget. The organized view additionally provides descriptive file paths and
metadata, links between related artifacts, and an index organized by evidence
role. A role identifies the type of evidence sought, such as a validation report
or the approved source code. Flat replaces the descriptive paths with numbered
item identifiers and removes the explicit structure; its underlying file
contents remain searchable and readable. These visible paths or item identifiers
serve as locators for accessing artifacts.
The final-only view exposes the terminal candidate and final summary artifacts,
omitting intermediate process and governance evidence.

The auditor is asked to locate each requested item of evidence and return its
locator and visible-content hash. It also reports findings about the first
process failure, if any, the Judge proposal, authorization and seal status,
and the identities of the approved, live, and frozen candidates.
The auditor uses GPT-5.5 at temperature 0.0 with up to twelve evidence-tool
calls per record--view audit. All views support listing, content search, and
reading by locator. Only the organized view supports batch lookup by evidence
role, returning locators and hashes for multiple requested roles in a single
budgeted call. Access is restricted to the assigned view, without preassigned
outcome-category labels or access to the original run directories.
The organized--flat comparison therefore evaluates structured
evidence organization together with its retrieval interface under this call
budget; comparison with final-only additionally examines evidence retention.

To compare how well the views support evidence recovery, we report Exact audit
evidence, the fraction of required audit roles that the auditor recovers
accurately. A role counts as accurately recovered only when the returned
artifact locator is correct and the hash matches its visible content. For
each record, the score is the number of accurately recovered required roles
divided by the number of audit roles required for that source record. The
reported value is the unweighted mean of these fractions over all 44 records
in a view.

For this metric, twelve evidence roles are required for every record: process
trace, Runtime Card, validation report, backtest report, Judge report, Judge proposal,
commitment seal, seal-verification result, governance events, run metric,
accepted code, and live code. Frozen hashes form a thirteenth required role
when the source record contains a frozen candidate; contained records without
one do not require this role. Any role required for a source record remains
required in all three views, even if a view does not expose it. Recovering
these audit roles does not by itself establish that the recorded outcomes can
be reproduced. The auditor also searches for additional inputs, including
task specifications, data bindings, and execution configurations, to support
the validation, backtesting, and Commitment reconstruction described next.

\subsection{Replay Inputs, Execution, and Reconstruction}
\label{app:audit-replay-protocol}

Replay uses the retrieved evidence to check two linked outcomes: validation
and backtesting of the approved candidate, and authorization of the candidate
presented for Commitment. The acceptance component requires ten roles:
task contract, strategy specification, blueprint, and implementation plan
describe the task and design; data binding, field contract, implementation
metadata, and accepted code identify the candidate and its data constraints;
validation and backtest configurations specify execution. After resolving
locators and verifying hashes, the runner restores these inputs in a fresh
run directory, reruns validation and, if it passes, the configured backtest
using the separately configured frozen market-data snapshot. It compares the
content hashes of the candidate code, data binding, and field contract with
those in the source record. The comparison also covers the validation outcome,
backtest status, and output protocol.

The Commitment component instead reconstructs the authorization outcome for
the live candidate at commitment time. It resolves live code, the saved Judge
proposal, commitment seal, seal verification, governance events, run metric,
and frozen hashes where a frozen candidate exists. It recomputes the live-code
identity and compares it with the seal, while using recorded verification
evidence for the supporting artifacts covered by the seal. The reconstruction checks authorization
and terminal candidate identity under the recorded enforcement condition.
The saved Judge proposal is an input; Judge review itself is not rerun.

Each component can proceed only with a complete input packet: all of its
required artifacts must be available through the retrieved locators with
matching visible-content hashes. The replay input interface accepts either
descriptive file paths or numbered item identifiers and restores the required
inputs in the same canonical workspace. All views use the same completeness
and execution checks.

A replay is verified only when both components match the source record's
specified outcomes and candidate identities. This includes
reproducing normal authorization and freezing for clean records, denied
authorization without freezing for contained records, and unsupported
freezing under Off for escaped records. Verification concerns these execution
and authorization outcomes and artifact identities; it does not require
byte-identical trading outputs or identical financial metrics.

\begin{table}[htbp]
  \caption{Audit and replay decomposition for 44 records per view. Packet
  completeness requires all roles for that component; execution counts are
  record counts, not average role-coverage percentages.}
  \label{tab:replay-decomposition}
  \centering\small
  \setlength{\tabcolsep}{4pt}
  \begin{tabular}{lrrrrrr}
    \toprule
    View & \shortstack{Exact audit\\evidence}
    & \shortstack{Acceptance\\packet}
    & \shortstack{Commitment\\packet}
    & \shortstack{Acceptance\\executed}
    & \shortstack{Commitment\\executed}
    & \shortstack{Verified\\replay} \\
    \midrule
    Organized & 100.0\% & 44/44 & 44/44 & 44/44 & 44/44 & 44/44 \\
Flat & 43.2\% & 0/44 & 6/44 & 0/44 & 6/44 & 0/44 \\
Final only & 27.6\% & 0/44 & 0/44 & 0/44 & 0/44 & 0/44 \\
\bottomrule

  \end{tabular}
\end{table}

The results in Table~\ref{tab:replay-decomposition} show why partial evidence
recovery need not yield a verified replay.
Flat recovers 43.2\% of audit evidence and completes
six Commitment sub-replays, but none of its 44 acceptance packets is complete.
Acceptance replay therefore cannot proceed, leaving zero verified replays.
Final-only recovers 27.6\% of audit evidence and completes neither component.
By contrast, the organized view completes both components and verifies
replay for all 44 records.

\section{FinResearchSuite}
\label{app:finresearchsuite}

\subsection{Task Composition and Interfaces}
\label{app:finresearchsuite-inventory}

FinResearchSuite complements FinRelBench by assessing research-task completion
and financial utility; these results are not pooled into the FinRelBench
reliability score. It contains 22 auxiliary tasks in three groups: open-ended
rule research, specified model development, and open-ended model research.
The groups differ in whether the strategies are rule- or model-based and in
how much of the design is specified by the public request. They are evaluated
separately.

Across the three groups, each public case contains an identifier, a
natural-language query, and a required output type. Daily scores support stock
ranking and portfolio construction, whereas intraday orders specify
timestamped buy and sell instructions.

\paragraph{RuleResearchBench (12 open-ended rule cases).}
Each request fixes a research objective and public constraints while leaving
the detailed signal construction, parameter choices, and portfolio design to
the system within those constraints.
The ten daily-score cases cover trend--value blending, market-regime rotation,
liquidity-quality selection, reversal with risk controls, low-volatility and
momentum mixing, price-limit-aware alpha, industry-balanced alpha,
gap--volatility alpha, cash-gated relative strength, and factor-bucket
ensembling. Two additional cases require intraday orders for liquidity reversal
and breakout-or-reversion research.

\paragraph{ModelBench (5 explicit model cases).}
Each case specifies the model to implement, with the five cases covering
LightGBM, gradient-boosted decision trees, a multilayer perceptron, a long
short-term memory network, and a Transformer encoder. The public requests
also specify input features or feature groups, prediction targets,
chronological train/validation/test separation, and daily-score output;
some also fix sequence lengths, architecture settings, or portfolio
constraints.

\paragraph{ModelResearchBench (5 open-ended model cases).}
In contrast to ModelBench, these requests set a research direction and any
model-family requirements while leaving more of the design to the system.
Within the public constraints, the system selects the model or its
configuration, feature set, preprocessing, and training design. The five cases
cover tabular, tree/boosting, neural, sequential, and Transformer-style alpha
research. All five cases require daily-score output.

\subsection{Evaluation Protocol}
\label{app:finresearchsuite-protocol}

The direct comparison includes QuantHarness, Codex CLI, OpenHands, SWE-agent,
and Aider. Human input is limited to the initial request. Each system then
follows its native agent loop without manual correction or strategy refinement.
All submitted strategies are evaluated by the shared backtester using the same
market data, trading costs, and execution assumptions within each task.

RD-Agent(Q) is additionally evaluated on the daily rule-research tasks. Its
factor outputs require a deterministic adapter to produce the daily scores
consumed by the shared backtester, so these results are reported separately.
The adapter does not produce intraday orders.

To separate candidate development from final performance assessment, the model
tasks use three non-overlapping data periods in chronological order: training,
optimization, and final evaluation. During candidate development, models are
fitted on the training period. Limited performance feedback from the
optimization period supports comparison and selection of candidates. The
selected design is then held fixed for a single evaluation on the final
period, which is used neither for fitting nor for candidate selection. The
final-period result is not returned for further strategy revision. For the
rule-research tasks, final evaluation likewise uses the selected strategy
implementation under the common backtest protocol.
Table~\ref{tab:research-windows} gives the dates and CSI~500 constituent
snapshots for these windows. Shared market and execution assumptions are
specified in Appendix~\ref{app:market-protocol}.

\begin{table*}[t]
  \caption{Chronological windows for the open-ended rule and model research
  protocol. Dates in the Universe column identify the constituent snapshots;
  each snapshot predates its evaluation or training window.}
  \label{tab:research-windows}
  \centering
  \small
  \setlength{\tabcolsep}{7pt}
  \begin{tabular}{llll}
    \toprule
    Phase & Dates & Universe & Use \\
    \midrule
    Training & 2025-01-02--2025-08-29 & CSI~500 (2024-12-31) & Fitting/construction \\
    Optimization & 2025-09-01--2025-12-31 & CSI~500 (2025-08-29) & Candidate selection \\
    Final & 2026-01-05--2026-05-29 & CSI~500 (2025-12-31) & One-time reporting \\
    \bottomrule
  \end{tabular}
\end{table*}
\FloatBarrier

FinResearchSuite uses GPT-5.5 for the five direct systems, with temperature
0.0 where supported and a six-hour runner time limit per task. The external
system releases are those listed in Table~\ref{tab:system-freeze}; the
FinRelBench time limits in Table~\ref{tab:reliability-budgets} do not apply
to these research runs. QuantHarness uses a 20-turn limit per agent invocation.

For both model groups, the published evaluation interface permits at most
three optimization evaluations and one final submission per system--task
pair. External systems perform their own model construction, training, and
inference; these evaluation limits do not impose QuantHarness's internal
proposal or ranking procedure on them. For rule research, QuantHarness uses
the bounded optimization route in Appendix~\ref{app:strategy-optimization},
while external systems produce their submissions through their native
workflows.

The common backtest protocol provides a shared basis for assessing the
different designs that can satisfy an open-ended request, without requiring a
unique reference implementation. Across all three groups, completion records
whether a strategy produces a final-period backtest, while financial quality
describes its performance in that backtest.
Appendix~\ref{app:finresearchsuite-metrics} defines these outcomes and their
aggregation.

\subsection{Completion and Financial Metrics}
\label{app:finresearchsuite-metrics}

A task is counted as completed only when it produces a final-period backtest.
For each system, completion uses all planned tasks as its denominator: 12 for
open-ended rule research and five for each model group. The rule group is
further separated into ten daily-score tasks and two intraday-order tasks,
with completion reported separately for these two output types.

Financial summaries cover the daily rule tasks and the two model groups;
intraday-order tasks are reported only through completion counts. For each
system, financial metrics are computed per completed task and averaged with
equal task weight within the corresponding group. Tasks without a final-period
backtest remain in the completion denominator but are excluded from financial
means, rather than being assigned zero returns. A group with no completed
tasks has no financial estimate. Financial means are therefore reported
alongside completion counts to show how many tasks contribute to them.

Daily rule-research results report annualized return, Sharpe ratio, maximum
drawdown, and turnover. Model results report total return, Sharpe ratio,
maximum drawdown, and the \emph{generalization gap} defined below.
The annualized returns of the daily rule tasks are not pooled with the total
returns of the model tasks. The
positive-return count in the model results reports how many completed tasks
earn a strictly positive final-period total return, with the number of
completed tasks as its denominator.

For each backtest, let $T$ be the number of recorded daily observations and
$V_t$ the total account equity at observation $t$, with $t=1,\ldots,T$ in
chronological order. Total return is $V_T/V_1-1$, and annualized return is
$(V_T/V_1)^{252/T}-1$, using 252 trading days per year.

To compute volatility and the Sharpe ratio, we use daily returns $r_1=0$ and
$r_t=V_t/V_{t-1}-1$ for $t=2,\ldots,T$. Annualized volatility is the
population standard deviation of these $T$ daily returns multiplied by
$\sqrt{252}$. The shared backtester computes Sharpe as annualized return
divided by this volatility, without subtracting a risk-free return; Sharpe
is set to zero when volatility is zero.

Maximum drawdown measures the largest peak-to-trough loss in the recorded
equity series. It is the minimum of $V_t/\max_{1\leq u\leq t}V_u-1$ over
the recorded observations, where $u$ indexes observations up to and including
$t$, so the denominator is the highest equity recorded by that point. It is
reported as a non-positive return, so a value closer to zero denotes a
smaller loss from peak to trough. Trading activity is measured by turnover:
the total monetary value of executed purchases and sales, counting both
positively, divided by mean account equity over the evaluation period. This
is reported as a cumulative multiple, not a daily or annualized rate.

For the model tasks, the generalization gap compares the selected candidate's
total return in final evaluation with its total return during optimization.
Let $R_{\mathrm{final}}$ and $R_{\mathrm{optimization}}$ denote these two
returns, respectively. We compute the gap as
\[
  \Delta R = R_{\mathrm{final}} - R_{\mathrm{optimization}}.
\]
For a given task, a negative gap indicates a lower total return in the final
period than in the optimization period; a positive gap indicates a higher
total return. For each system, these task-level gaps are averaged with equal
weight over completed tasks within each model group and reported in percentage
points. The gap compares returns between the two periods; it does not measure
performance across other market periods or regimes.

\subsection{QuantHarness Optimization and Candidate Selection}
\label{app:strategy-optimization}

This subsection details how QuantHarness develops and selects candidates for
the final evaluation in FinResearchSuite. The optimization and selection
rules below are specific to QuantHarness; the shared evaluation protocol is
described in Appendix~\ref{app:finresearchsuite-protocol}.

Passing validation and backtesting establishes that a candidate can be executed
with the bound inputs and declared output protocol. At this point, QuantHarness
supports an optional, budget-limited exploration of whether its financial quality can
be improved while preserving the task requirements. The Workflow can complete
without this loop.
During optimization, the task requirements, data binding, and output protocol
remain unchanged. Repair addresses execution or interface errors; optimization
explores improvements to an executable candidate, and the two budgets are
not interchangeable.

\paragraph{Protocol and candidate proposals.}
The optimization budget and financial-quality objective are fixed before the
run. In FinResearchSuite, QuantHarness allows at most three optimization
evaluations and one final submission per task. Market data are divided into
non-overlapping optimization and final evaluation periods; during candidate
development, model-based strategies fit
on an earlier training period.

Within this protocol, the proposing agent receives the permitted changes and
evaluation settings, the current candidate, and optimization-period feedback,
not final-period results.
A proposal must name an allowed change and remain consistent with the task,
data binding, and output protocol. The proposal guard checks whether the
requested change is permitted before development; the variant guard then
checks the resulting candidate against the proposal's constraints and the
fixed task and data requirements. Model-based variants also produce the
training and inference artifacts required for validation and backtesting.
Each variant must pass validation before backtesting on the same optimization
period.

\paragraph{Budget accounting and evidence.}
To keep exploration within the declared budget, the Workflow tracks candidate
identities, proposal iterations, optimization evaluations, repairs, final
submissions, and final evaluation attempts separately.
An append-only budget ledger records evaluation starts and completions,
candidate identity, and submission status. Starting an evaluation consumes
its applicable allowance, even if the attempt subsequently fails.

To identify duplicate candidates, hashes are computed from the designated
files that define the strategy and its implementation, including model
configuration where applicable. A duplicate candidate does not receive a new
independent evaluation. Instead, it may reuse a recorded result only if that
result comes from a completed evaluation of the same candidate under a
matching recorded execution protocol. Incomplete or incompatible cached
evidence cannot make a candidate eligible for selection.

\paragraph{Stopping and selection.}
Exploration ends and the Workflow proceeds to selection when the configured
iteration, candidate, or evaluation budget is exhausted, or when no actionable
improvement or further admissible proposal is available. Rejection of a new
proposal, for example because it attempts to change the fixed data environment,
blocks that proposal and can end exploration without invalidating previously
eligible candidates.

By contrast, a run-level protocol violation, such as attempting to evaluate
an unregistered candidate or replace an already registered selection with
another candidate, terminates the run without a valid final submission.
When the protocol remains valid, selection considers both the initial candidate
and its revisions. Eligibility requires the applicable validation and backtest
checks, not just a favorable financial number. If no eligible candidate remains,
the run terminates without a valid final submission. Otherwise, the eligible
candidates are compared under the fixed objective using optimization-period
evidence. The selected candidate's identity, the ranking, and the record of
its final submission are retained.

In FinResearchSuite, QuantHarness uses a composite financial score rather than
return alone. For the eligible candidates in the reported runs, the score
adds annualized return and 0.5 times the Sharpe ratio, then subtracts 0.5 times
absolute maximum drawdown and 0.05 times cumulative turnover. Returns and
drawdowns enter as fractions; the metric conventions are defined in
Appendix~\ref{app:finresearchsuite-metrics}. Rule-task scoring adds 0.01 per
executed fill, capped at 0.1. Model-task scoring halves this fill-count bonus
and adds 0.05 times the fill rate, the ratio of executed to requested shares,
clipped to $[0,1]$. Turnover above 2 incurs a further penalty of 0.25.

Rule-research candidates are ranked by this score over the optimization
period. Model candidates instead require complete evidence for four monthly
subperiods within that window, September--December 2025, with at least ten daily
observations per subperiod. Their ranking score is the median monthly score
minus half the population standard deviation of the four scores and half the
difference between the median and worst monthly scores. A model variant must
also outperform the candidate it revises on at least three of the four monthly
scores to remain eligible. Selection
takes the highest-ranked eligible candidate; equal scores retain candidate
registration order. These calculations use only optimization-period evidence.

\paragraph{Final evaluation and engineering acceptance.}
After selection, the candidate's design is held fixed for a single evaluation
on the final period. For QuantHarness's model tasks in FinResearchSuite,
preparation for this evaluation includes refitting the selected model on the
training and optimization periods and generating predictions for the final
period, without revising the selected design.
The final attempt is registered before execution; failure does not authorize
reselection or a second final attempt.
Final financial results are retained for quality analysis and are not returned
to optimization or used to retune the strategy. Their magnitude does not
determine engineering acceptance.

The candidate then proceeds to Judge review
with the corresponding validation and backtest reports. Harness Core checks
the evidence supporting the Judge's recommended route before allowing it to
take effect. Formal Workflow freezing follows an authorized acceptance
decision and preserves the candidate together with its supporting artifacts;
holding a candidate fixed for evaluation does not itself constitute freezing.
Experimental windows and generation budgets are specified in
Appendix~\ref{app:finresearchsuite-protocol}.

Full financial results and their interpretation are provided in
Appendix~\ref{app:e3-results}.

\section{FinResearchSuite Results}
\label{app:e3-results}

This appendix reports the auxiliary evaluation results on FinResearchSuite. Task definitions and the evaluation 
protocol appear in Appendix~\ref{app:finresearchsuite}. Completion is measured
over all planned tasks, while financial means include only completed
final-period backtests. We retain the three task groups separately so that
differences in their specification, output type, and completion coverage remain
visible.

\subsection{Daily Open-Ended Rule Research}

The ten daily-score tasks test whether systems can turn an open-ended research
idea into a strategy that can be evaluated under the common market protocol.
Human involvement ends after the initial request, and all systems are evaluated
under the same cost and slippage rules within each task, as specified in
Appendix~\ref{app:market-protocol}.

Table~\ref{tab:openresearch-daily} reports completion and the financial performance
of the resulting strategies, using the metrics defined in
Appendix~\ref{app:finresearchsuite-metrics}. RD-Agent(Q) is shown separately
because its factor outputs require an adapter to produce daily scores for the
shared backtester.

\begin{table}[!htbp]
  \caption{Financial results for the ten daily-score tasks in
  RuleResearchBench. Financial metrics are means over each system's completed
  tasks.}
  \label{tab:openresearch-daily}
  \centering
  \small
  \setlength{\tabcolsep}{4pt}
  \renewcommand{\arraystretch}{1.05}
  \begin{tabular*}{\textwidth}{@{\extracolsep{\fill}}lrrrrr@{}}
    \toprule
    System
      & \shortstack{Completed\\tasks $\uparrow$}
      & \shortstack{Mean ann.\\return $\uparrow$}
      & \shortstack{Mean\\Sharpe $\uparrow$}
      & \shortstack{Mean max\\drawdown $\uparrow$}
      & \shortstack{Mean\\turnover $\downarrow$} \\
    \midrule
    QuantHarness & \textbf{10/10} & $\mathbf{-3.27\%}$ & $\mathbf{-0.219}$
      & $\mathbf{-13.85\%}$ & \textbf{33.06} \\
    Codex CLI & \textbf{10/10} & $-10.65\%$ & $-0.611$ & $-15.03\%$ & 51.63 \\
    OpenHands & \textbf{10/10} & $-5.68\%$ & $-0.350$ & $-14.95\%$ & 59.75 \\
    SWE-agent & 9/10 & $-13.07\%$ & $-0.732$ & $-14.50\%$ & 59.90 \\
    Aider & 9/10 & $-5.59\%$ & $-0.384$ & $-15.22\%$ & 62.26 \\
    \addlinespace[2pt]
    RD-Agent(Q) & \textbf{10/10} & $-7.66\%$ & $-0.503$ & $-18.75\%$ & 57.31 \\
    \bottomrule
  \end{tabular*}
  \vspace{1pt}
  \parbox{\textwidth}{\footnotesize
    \textit{Notes.}
    Missing tasks are not imputed as zero returns. Turnover is a cumulative
    multiple. RD-Agent(Q) uses a factor-to-score adapter for the daily
    interface. Bold marks the best observed value in each column, including ties.}
\end{table}

QuantHarness, Codex CLI, OpenHands, and RD-Agent(Q) complete all ten tasks;
SWE-agent and Aider complete nine each. Among the completed strategies,
QuantHarness has the highest mean annualized return and Sharpe ratio, the
smallest mean drawdown magnitude, and the lowest mean turnover. Mean return
and Sharpe remain negative for all systems, so this is a relative advantage
among the completed strategies under the shared evaluation conditions.
\FloatBarrier

\subsection{Specified and Open-Ended Model Development}

The two model groups assess completion and financial performance under
different degrees of research freedom. ModelBench provides more explicit
modeling requirements, whereas ModelResearchBench leaves more choices about
features, preprocessing, and training to the system within the public
request's constraints.
Table~\ref{tab:model-results} reports their completion and final-period
financial results separately.

\begin{table}[!htbp]
  \caption{Financial results for ModelBench and ModelResearchBench. Financial
  metrics are means over each system's completed tasks within the corresponding
  group.}
  \label{tab:model-results}
  \centering
  \small
  \setlength{\tabcolsep}{3.5pt}
  \renewcommand{\arraystretch}{1.07}
  \begin{tabular*}{\textwidth}{@{\extracolsep{\fill}}lrrrrrr@{}}
    \toprule
    System
      & \shortstack{Completed\\tasks $\uparrow$}
      & \shortstack{Positive\\returns $\uparrow$}
      & \shortstack{Mean total\\return $\uparrow$}
      & \shortstack{Mean\\Sharpe $\uparrow$}
      & \shortstack{Mean max\\drawdown $\uparrow$}
      & \shortstack{Mean generalization\\gap (pp)} \\
    \midrule
    \multicolumn{7}{l}{\textit{A. ModelBench-5: specified model families}} \\
    QuantHarness & \textbf{5/5} & \textbf{3/5} & $\mathbf{+1.75\%}$
      & $\mathbf{0.168}$ & $\mathbf{-15.32\%}$ & $+4.32$ \\
    Codex CLI & \textbf{5/5} & 2/5 & $-0.77\%$ & $-0.170$ & $-17.27\%$ & $-0.82$ \\
    OpenHands & \textbf{5/5} & 2/5 & $+1.24\%$ & $0.058$ & $-16.05\%$ & $+0.83$ \\
    SWE-agent & 3/5 & 0/3 & $-14.79\%$ & $-1.152$ & $-22.63\%$ & $-15.53$ \\
    Aider & 0/5 & -- & -- & -- & -- & -- \\
    \addlinespace[3pt]
    \multicolumn{7}{l}{\textit{B. ModelResearchBench-5: open-ended model research}} \\
    QuantHarness & \textbf{5/5} & \textbf{4/5} & $\mathbf{+1.29\%}$
      & $\mathbf{0.132}$ & $\mathbf{-15.93\%}$ & $+2.39$ \\
    Codex CLI & \textbf{5/5} & 2/5 & $-5.64\%$ & $-0.748$ & $-18.41\%$ & $-6.50$ \\
    OpenHands & \textbf{5/5} & 1/5 & $-5.58\%$ & $-0.617$ & $-16.86\%$ & $-8.05$ \\
    SWE-agent & 0/5 & -- & -- & -- & -- & -- \\
    Aider & 0/5 & -- & -- & -- & -- & -- \\
    \bottomrule
  \end{tabular*}
  \vspace{1pt}
  \parbox{\textwidth}{\footnotesize
    \textit{Notes.}
    The positive-return count uses completed tasks as its denominator.
    Missing tasks are not imputed as zero returns. Generalization gaps are in
    percentage points (pp). Bold marks the best observed
    completion, positive-return count, and final-period financial metrics in
    each panel, including ties.}
\end{table}

QuantHarness, Codex CLI, and OpenHands complete all five tasks in each model
group. SWE-agent completes three specified-model tasks and no open-ended model
tasks; Aider does not complete any task in either group.

QuantHarness is the only system with positive mean total return and Sharpe
ratio in both groups. Three of its five
specified-model strategies and four of its five open-ended model strategies
earn positive final-period returns. The open-ended model group illustrates why
completion and financial quality must be considered separately: Codex CLI and
OpenHands also complete every task, but both have negative mean final-period
returns.
Thus, a completed modeling pipeline does not by itself establish the utility
of the selected strategy.

To compare the selected candidates' final-period returns with their
optimization-period returns, we also report the generalization gap
(Appendix~\ref{app:finresearchsuite-metrics}).
QuantHarness's mean generalization gaps are $+4.32$ and $+2.39$ percentage
points for the specified and open-ended model groups, respectively. These
values indicate higher mean total returns in the final period than in the
optimization period for the selected candidates in each group. Codex CLI and
OpenHands instead have negative mean gaps in the open-ended model group,
indicating lower final-period returns than optimization-period returns on
average.

\subsection{Intraday Output Coverage}

Beyond the daily-score tasks above, the two intraday rule-research tasks
require timestamped buy and sell orders. Here, we assess whether systems can
also complete research tasks through this order-based interface and execution
path. We therefore report completion coverage for these two tasks; the
financial comparisons above remain scoped to daily-score strategies.

\begin{table}[htbp]
  \caption{Completion of the two intraday rule-research tasks.}
  \label{tab:finresearch-intraday-completion}
  \centering
  \small
  \begin{tabular}{lr}
    \toprule
    System & Completed tasks \\
    \midrule
    QuantHarness & \textbf{2/2} \\
    Codex CLI & 0/2 \\
    OpenHands & 1/2 \\
    SWE-agent & 1/2 \\
    Aider & 0/2 \\
    \midrule
    Total (system--task runs) & 4/10 \\
    \bottomrule
  \end{tabular}
\end{table}

Across the five direct systems, four of ten system--task runs produce a
final-period backtest (Table~\ref{tab:finresearch-intraday-completion}).
QuantHarness is the only system to complete both tasks, extending its
completion coverage from daily-score research to the intraday-order interface.
RD-Agent(Q) is excluded because the adapter used here produces daily scores
rather than intraday orders.

\subsection{Scope of the Auxiliary Evidence}

QuantHarness completes all 22 tasks under the common final-period evaluation
and obtains favorable relative financial results.
The profitable model-based candidates provide evidence of practical research
utility, while the negative average daily rule returns show that successful
development need not produce a profitable strategy. Together, these results
complement the reliability experiments by showing that the governed Workflow
can carry less constrained strategy research through to final-period evaluation.

For this task set and final period, financial means should be read alongside
completion coverage. These financial rankings compare complete
systems; the paired boundary tests in Appendix~\ref{app:e2-results} separately
examine the contribution of individual governance boundaries to runtime
reliability.

\section{Implementation and Reproducibility Details}
\label{app:implementation}
\label{app:method}
\label{app:reproducibility}

This appendix connects the method to its operational interfaces. It identifies
stage-level decision authority and follows a real candidate through development
and freezing, before giving the experimental settings and reproducibility scope.

\subsection{Stage Responsibilities and Decision Authority}
\label{app:stage-controls}

The Workflow defines the finance-specific operations at each stage. Harness
Core controls which information a stage receives, which tools it may invoke,
and whether the run may change state on the basis of its outputs.
Table~\ref{tab:stage-controls} summarizes the inputs,
permitted actions, required evidence, and decision authority at each stage;
the following paragraphs explain how the three boundaries apply across these
stages. The optional optimization route is detailed separately in
Appendix~\ref{app:strategy-optimization}.

\begin{table}[!ht]
  \caption{Stage interfaces and authority. An agent proposal or a written
  report is not itself permission to advance the run.}
  \label{tab:stage-controls}
  \small
  \begin{tabularx}{\textwidth}{@{}>{\raggedright\arraybackslash}p{0.16\textwidth}XX@{}}
    \toprule
    Stage & Stage inputs and permitted actions & Required evidence and authority \\
    \midrule
    Task understanding &
    Public request and operating mode; structured extraction of intent,
    missing fields, conflicts, and assumptions; no strategy-code generation. &
    Task contract. Intake policy determines whether the task is ready to
    proceed or needs clarification; the runtime checks the associated
    state transition. \\
    \addlinespace
    Planning and data grounding &
    Public task and stage-specific Runtime Card. Catalog listing and asset
    inspection precede binding; only authorized data tools are exposed. &
    Specification, planning blueprint, data binding, and field contract.
    Deterministic catalog/schema checks verify the proposed sources and
    fields. \\
    \addlinespace
    Construction &
    Accepted specification, bound fields, implementation plan, and required
    output interface. Code generation uses the established data binding
    rather than choosing new data sources. &
    Candidate source and implementation metadata; model paths additionally
    produce training/inference artifacts. Stages pass candidates to checks,
    not directly to acceptance. \\
    \addlinespace
    Validation and backtest &
    Deterministic services validate and backtest the candidate under the
    configured data and execution protocol; they do not redesign the strategy. &
    Validation checks, execution outputs, and backtest reports. Failed checks
    block normal progression; recovery requires authorization. \\
    \addlinespace
    Recovery &
    Public failure feedback and the current candidate, with stage-limited
    repair/replan/rebind authority and bounded attempts. &
    Proposed change, guard result, revised candidate, and renewed checks.
    Recovery cannot silently weaken the task to make execution pass. \\
    \addlinespace
    Judge and commitment &
    Judge reads a compact evidence bundle and recommends acceptance,
    recovery, or stopping; it does not execute its own recommendation. &
    Judge report, policy/guard decision, and approved candidate identity.
    The guarded runtime authorizes transitions. Before saving the frozen
    artifact, the freezer checks required evidence and, when seal enforcement
    is enabled, verifies that the files match their approved versions. \\
    \bottomrule
  \end{tabularx}
\end{table}

\paragraph{Information boundary.}
For each agent-facing stage, this boundary separates the context supplied to
the model from information retained for governance and audit.
The Runtime Card stores \texttt{agent\_visible} and
\texttt{harness\_private} fields. The former contains the current goal, task
intent, public assumptions and feedback, constraints, allowed tools, and
required outputs. The latter records context budgets, policy notes, which
information is excluded from the request, and references to source artifacts.
The complete card is retained for audit; each stage's request builder uses
the agent-visible fields to assemble the model messages. The information
boundary is therefore applied when the stage's request is assembled, rather
than by passing the complete
audit record to the agent.

\paragraph{Action boundary.}
The allowed-tool list in the Runtime Card is enforced both when tools are
exposed to the agent and when a call is invoked. Tool definitions and their
execution handlers are registered once, but only the stage's permitted tools
are exposed. Before executing a call, the runtime checks permission again.
If the call is denied, the runtime records the requested tool, arguments,
allowed-tool list, and rejection before returning a tool error.
Appendix~\ref{app:normal-boundary-tests} describes how the Action ablation
changes tool exposure, guidance, and enforcement.

\paragraph{Commitment boundary.}
Producing candidate files or reports does not itself authorize the run to
change state. The guarded runtime checks whether the proposed transition is
allowed and whether its required evidence is available. At candidate review,
the Judge recommends acceptance, recovery, or stopping; it does not itself
change the run's state. When the authorized route leads to freezing, the freezer checks
Judge acceptance, validation, backtest, and required artifacts before copying
key files and recording their hashes. The runtime, rather than the freezer,
records the run's completion after freezing succeeds.

In configurations with seal enforcement, freezing additionally checks that
the candidate and supporting evidence still match the approved versions.
A \emph{commitment seal} records the paths and content hashes of those files;
the freezer checks the current files against this record before commitment.
Financial performance remains separate from these engineering prerequisites.
The paired ablations and controlled interventions in
Appendices~\ref{app:normal-boundary-tests} and~\ref{app:boundary-stress-tests}
evaluate these boundaries.
\FloatBarrier

\subsection{A Complete Development Trajectory from Request to Freeze}
\label{app:implementation-example}

To illustrate how these stage responsibilities connect in a recorded run, we
follow a completed Full QuantHarness trajectory for the daily moving-average
crossover task,
selected for its explicit request and complete development records.
The run uses a rule-based internal Judge, one backtest attempt, and no repair
or optional optimization.

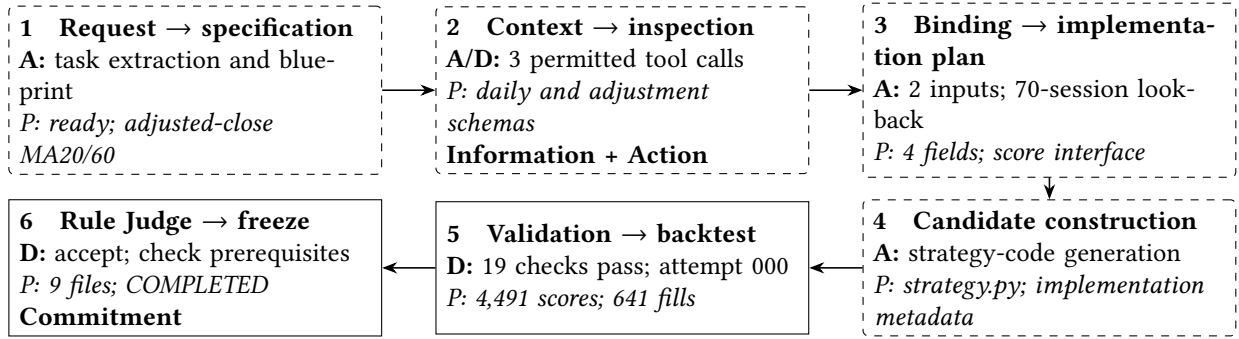
\begin{figure}[htbp]
  \centering
  \begin{tikzpicture}[
  font=\fontsize{10}{12}\selectfont,
  stage/.style={draw,align=left,text width=4.65cm,minimum height=1.75cm,
    inner sep=4pt},
  proposed/.style={stage,dashed,rounded corners=2pt},
  checked/.style={stage,solid},
  flow/.style={-{Stealth[length=2mm]},line width=0.6pt},
]
\node[proposed] (one) at (0,0) {
  \textbf{1\quad Request $\rightarrow$ specification}\\
  \textbf{A:} task extraction and blueprint\\
  \textit{P: ready; adjusted-close MA20/60}
};
\node[proposed] (two) at (5.65,0) {
  \textbf{2\quad Context $\rightarrow$ inspection}\\
  \textbf{A/D:} 3 permitted tool calls\\
  \textit{P: daily and adjustment schemas}\\
  \textbf{Information + Action}
};
\node[proposed] (three) at (11.3,0) {
  \textbf{3\quad Binding $\rightarrow$ implementation plan}\\
  \textbf{A:} 2 inputs; 70-session lookback\\
  \textit{P: 4 fields; score interface}
};
\node[proposed] (four) at (11.3,-2.35) {
  \textbf{4\quad Candidate construction}\\
  \textbf{A:} strategy-code generation\\
  \textit{P: strategy.py; implementation metadata}
};
\node[checked] (five) at (5.65,-2.35) {
  \textbf{5\quad Validation $\rightarrow$ backtest}\\
  \textbf{D:} 19 checks pass; attempt 000\\
  \textit{P: 4,491 scores; 641 fills}
};
\node[checked] (six) at (0,-2.35) {
  \textbf{6\quad Rule Judge $\rightarrow$ freeze}\\
  \textbf{D:} accept; check prerequisites\\
  \textit{P: 9 files; COMPLETED}\\
  \textbf{Commitment}
};
\draw[flow] (one.east) -- (two.west);
\draw[flow] (two.east) -- (three.west);
\draw[flow] (three.south) -- (four.north);
\draw[flow] (four.west) -- (five.east);
\draw[flow] (five.west) -- (six.east);
\end{tikzpicture}
  \caption{The observed request-to-freeze trajectory. Numbers match the
  content boxes below. A denotes agent proposals/generation, D deterministic
  operations, and P persisted evidence. Dashed rounded nodes contain agent
  stages; solid nodes contain deterministic review and execution. The three
  boundary annotations identify the mediation observed in this run.}
  \label{fig:request-to-freeze}
\end{figure}

Figure~\ref{fig:request-to-freeze} gives an overview; Boxes~1--6 follow the
same candidate through the requests, decisions, and saved artifacts at each step.
The original request is quoted verbatim; other excerpts are shortened
or translated, code omissions are marked, and displayed numerical
samples are rounded. Code is rewrapped without changing its tokens.

\FloatBarrier

\begin{trajectorybox}{1}{Request and accepted specification}
\textbf{Original request.}
Implement a daily moving-average crossover strategy using adjusted close
prices. Compute 20-day and 60-day moving averages. Generate positive scores
when the short average is above the long average, rebalance on the next
trading session, hold at most 20 names, equal weight positions, and include
transaction costs and slippage.

\smallskip
\textbf{Task Contract.}
\texttt{mode=benchmark; status=ready; conflicts=[]}.
The contract records eight missing items: universe, backtest period, cost
rate/model, slippage rate/model, ranking beyond 20 eligible names, exact score
formula, cash handling, and long/short constraints. Here, the benchmark-mode
policy allows development to proceed with recorded defaults and assumptions,
including a long-only interpretation, rather than treating these details as
supplied by the user.

\smallskip
\textbf{Specification and Blueprint.}
For each security and signal date, MA20 and MA60 denote the simple means
of the latest 20 and 60 daily adjusted-close observations available through
that date, respectively; both require complete windows. Eligibility requires
MA20 $>$ MA60; execution occurs in the next session, with at most 20
equal-weight targets. The Blueprint specifies
$\max(\mathrm{MA20}/\mathrm{MA60}-1,0)$ as the fallback when no benchmark
score formula is supplied. The ratio is evaluated only when both averages
are finite and MA60 is positive. The Blueprint also specifies descending-score
selection when more than 20 names qualify and no benchmark ranking rule is
supplied. Signal construction remains
separate from portfolio/execution settings.
\end{trajectorybox}
The ready contract and supported rule-based specification permit entry into
development. The binding agent must now identify available data sources and
fields for this plan without changing the requested strategy.

\begin{trajectorybox}{2}{Stage context and observed actions}
\textbf{Saved binding request, translated excerpt.}
\emph{Perform data binding only; do not write strategy.py or executable code.}
The supplied visible context has \texttt{current\_state=DEVELOPING} and
\texttt{task\_status=ready}, and requires matching the Blueprint without
changing the user's intent or inventing hidden configuration values.

\smallskip
\textbf{Exposed tools.}
\texttt{list\_data\_catalog}, \texttt{inspect\_data\_asset},
\texttt{inspect\_quant\_layer}, \texttt{build\_strategy\_dataset}.
The three observed calls, all completed, are catalog listing
(\texttt{layer=all, include\_files=false}), then inspection of the daily-price
asset (\texttt{clean:daily}) and adjustment-factor asset
(\texttt{clean:adj\_factor}).

\smallskip
\textbf{One actual call and observation.}
\texttt{inspect\_data\_asset} with
\texttt{asset\_id="clean:adj\_factor"}, \texttt{sample\_rows=5},
\texttt{include\_stats=false} returns \texttt{ok=true}, daily frequency, fields
\texttt{[ts\_code, trade\_date, adj\_factor]}, and coverage
2020-01-02--2026-05-29.
\end{trajectorybox}
These inspections identify the price and adjustment-factor sources needed
to derive adjusted close prices. The resulting data binding and field contract
specify how their fields will be presented to the generated code.

\begin{trajectorybox}{3}{Grounded data and implementation plan}
\begin{tabularx}{\linewidth}{@{}lX@{}}
  \toprule
  Binding or plan item & Recorded content \\
  \midrule
  Price input & \texttt{clean:daily.close} $\rightarrow$
    \texttt{daily\_\_close} \\
  Adjustment input & \texttt{clean:adj\_factor.adj\_factor} $\rightarrow$
    \texttt{adj\_factor\_\_adj\_factor} \\
  Alignment and history & Daily panel joined by security
    (\texttt{ts\_code}) and date (\texttt{trade\_date});
    70 trading sessions of lookback \\
  Field contract & The two keys and two bound value columns above;
    output \texttt{signal\_date, symbol, score} \\
  Implementation plan & Multiply close by adjustment factor; sort by
    security/date; compute full-window MA20/60; retain finite positive
    crossover scores \\
  Execution responsibility & Portfolio capacity, equal weighting,
    next-session execution, costs and slippage are delegated to the
    backtest framework, not hardcoded into the signal calculation \\
  \bottomrule
\end{tabularx}
\end{trajectorybox}
The field contract fixes the source-to-code mapping, while the plan specifies
the derived calculations. Together they guide code generation; the market
dataset is assembled later, after validation, as part of the backtest attempt.

\begin{trajectorybox}{4}{Candidate implementation}
\textbf{Source excerpts.}
The single candidate has source SHA-256 prefix \texttt{012e769de4b3},
used below to link the executed and frozen code. The excerpts implement
the adjusted-price calculation, moving averages, and scoring rule from Box~3.
The surrounding code checks required columns, sorts by security/date,
removes duplicate keys, and filters invalid numeric values.
\begin{lstlisting}[style=trajectorypython]
short_window = 20
long_window = 60
# ...
work["derived_adjusted_close"] = work["daily__close"] * work["adj_factor__adj_factor"]
# ...
work["derived_ma20"] = work.groupby("ts_code", sort=False)[
    "derived_adjusted_close"].transform(
    lambda series: series.rolling(
        window=self.short_window, min_periods=self.short_window).mean())
work["derived_ma60"] = work.groupby("ts_code", sort=False)[
    "derived_adjusted_close"].transform(
    lambda series: series.rolling(
        window=self.long_window, min_periods=self.long_window).mean())
# ...
work["derived_score"] = (work["derived_ma20"] / work["derived_ma60"]) - 1.0
# ...
valid_score = (
    pd.notna(work["derived_ma20"])
    & pd.notna(work["derived_ma60"])
    & pd.notna(work["derived_score"])
    & (work["derived_ma60"] > 0.0)
    & (work["derived_ma20"] > work["derived_ma60"])
    & (work["derived_score"] > 0.0)
)
\end{lstlisting}
\end{trajectorybox}
The Workflow next checks the candidate's syntax, interface, and data usage
before allowing it to execute in backtest attempt \texttt{000}.

\begingroup
\tcbset{breakable,title after break={5\quad Validation and execution evidence (continued)}}
\begin{trajectorybox}{5}{Validation and execution evidence}
\textbf{Validation report.} All 19 checks pass; errors = 0; warnings = 0.
Selected recorded checks:
\begin{center}
\begin{tabular}{@{}ll@{}}
  \toprule
  Check & Result \\
  \midrule
  Python syntax & passed \\
  Dataset specification & passed; two inspected assets \\
  Strategy interface & passed; \texttt{scores} \\
  Dataset column access & passed; four bound columns \\
  Lightweight execution (dry-run) & passed; 62 output rows \\
  \bottomrule
\end{tabular}
\end{center}
The dry-run checks that the candidate loads, executes, and returns the
required score fields; it is not the market backtest. With validation passed,
the validation gate permits entry into backtesting, recording
\texttt{status=approved; action=enter\_backtesting}.

\smallskip
\textbf{Backtest attempt 000.}
The runner assembles 16,481 daily input rows, including history from
2025-09-17, for a 100-name universe and an evaluation window of
2026-01-05--2026-05-29. Recorded execution settings specify session-close
signals and a one-session lag; effective portfolio settings are
\texttt{max\_names=20} and \texttt{weighting=equal}. Costs include a
0.0003 commission rate (minimum CNY~5) and 5-basis-point slippage.

\smallskip
\textbf{Generated scores.}
\begin{center}
\begin{tabular}{@{}lll@{}}
  \toprule
  \texttt{signal\_date} & \texttt{symbol} & \texttt{score} \\
  \midrule
  2026-01-05 & 000100.SZ & 0.043810 \\
  2026-01-05 & 000333.SZ & 0.028322 \\
  2026-01-05 & 000338.SZ & 0.059951 \\
  \bottomrule
\end{tabular}
\end{center}
The saved scores contain 4,491 rows, of which 4,124 lie in the requested
evaluation window; the remainder come from the supplied history.
The output manifest records the dataset identity and source hash
\texttt{012e769de4b3}, linking these scores to the inputs and code that
produced them.

\smallskip
\textbf{From scores to an observed order and fill.}
The backtest framework converts these scores into portfolio targets using
the recorded position limit and equal-weight setting.
The 2026-01-05 score for \texttt{000100.SZ} leads to a buy on 2026-01-06:
target weight 0.05; 10,400 shares requested and filled;
volume-weighted average price (VWAP) 4.795185;
fill price 4.797583 after slippage; fees 15.467407
(prices and fees in CNY).
The report records \texttt{ok=true}, 95 equity observations, 641 orders,
and 641 fills. The backtest evidence guard approves the evidence after
confirming that the latest and attempt-specific reports refer to the same
attempt and that the referenced output manifest is available.
\end{trajectorybox}
\endgroup
Thus adjusted-price inputs and 20/60 windows are visible in the binding and
code, while the next-session requirement has a dated execution example.
The internal Judge reviews this engineering evidence to decide whether the
candidate can be frozen. This Workflow decision is separate from the external
requirement-level evaluation defined in
Appendix~\ref{app:benchmark-details}.

\begin{trajectorybox}{6}{Judge decision and frozen artifact}
\textbf{Internal review.}
For this review, an analysis bundle collects the task, specification,
implementation metadata, validation and backtest reports, tool-use summary,
and repair count. The rule-based Judge accepts the candidate because
validation and backtesting pass and trades are recorded:
\begin{center}
\begin{tabular}{@{}ll@{}}
  \toprule
  Field & Value \\
  \midrule
  \texttt{judge\_mode}; \texttt{llm\_judge\_used} & \texttt{rule}; \texttt{false} \\
  \texttt{decision}; \texttt{accepted\_for\_freeze} & \texttt{freeze}; \texttt{true} \\
  \texttt{primary\_reason} & \nolinkurl{validation_and_backtest_passed_with_trades} \\
  \texttt{issues}; repair attempts & \texttt{[]}; 0 \\
  \bottomrule
\end{tabular}
\end{center}
\textbf{Authorization and persistence.}
The runtime guard authorizes the transition from candidate review
(\texttt{ANALYZING}) to \texttt{FREEZING}.
The recorded freeze-precondition check has no violations. Nine files
are copied: Task Contract, specification, Blueprint, implementation metadata,
source, validation report, backtest report, analysis bundle, and Judge report.

\smallskip
\textbf{Linking executed and frozen artifacts.}
The source hash in the execution output manifest matches the frozen source
hash \texttt{012e769de4b3}; all nine copied files match their indexed
SHA-256 values. The dataset, scores, and backtest tables remain linked
through reports and manifests rather than being copied into the frozen package.
After the frozen candidate and hash index exist, the final guard approves
the transition from \texttt{FREEZING} to \texttt{COMPLETED}; the state
history records \nolinkurl{candidate_frozen} as the reason for completion.
This run records freeze-time hashes but uses no separate pre-freeze
commitment seal. Seal enforcement against post-Judge candidate replacement
is evaluated separately in Appendix~\ref{app:boundary-stress-tests}.
\end{trajectorybox}
The trajectory connects the request to grounded inputs, executable rules,
dated trading evidence, and an authorized frozen artifact. The rule-based
Judge's acceptance rests on successful validation and backtesting with
recorded trades, not on a return threshold.

\subsection{Prompt Constraints and Deterministic Checks}
\label{app:prompt-constraints}

The preceding example follows one run from request to freeze. Here, we explain
the reusable instructions that guide agent proposals and the deterministic
checks that govern whether their outputs can advance the Workflow. The five
activities covered are task understanding, data binding, candidate construction,
repair, and LLM-assisted Judge review. They are not five mandatory steps:
repair requires an authorized recovery route, and the Judge instructions apply
only when LLM-assisted review is enabled.

The boxes present English translations of core instructions, retaining their
requirements and identifiers. Shared rules are printed once, with their
applicable activities indicated, followed by the activity-specific instructions.
This template-level presentation complements the saved binding request in
Appendix~\ref{app:implementation-example}; it is not a transcript of that run.
Run-specific context, tool definitions, full output schemas, and additional
detailed execution-policy and library-compatibility blocks are not expanded
here. Boxed text addresses the agent; the surrounding prose explains the
checks performed outside the model.

\begin{promptbox}{Shared prompt blocks --- applicability by activity}
\textbf{Output format --- all five activities below.}

Output-format constraint: output exactly one JSON object; do not output
Markdown, explanatory paragraphs, or code blocks.

\textbf{Backtest-result boundary --- task understanding, data binding,
candidate construction, and repair.}
\begin{promptitems}
  \item Do not run backtests or claim to have obtained returns, Sharpe ratios,
  maximum drawdown, or other results.
  \item You may only state that the current stage has produced a plan, binding,
  implementation, repair, or review recommendation.
\end{promptitems}

\textbf{Cross-stage requirement integrity --- data binding and candidate
construction.}
\begin{promptitems}
  \item \texttt{agent\_visible.raw\_request} (\texttt{task.raw\_request} during
  code generation) is the complete public request and the basis for requirements.
  Task summaries, StrategySpec, and implementation plans may refine it, but must
  not remove or override its requirements. Preserve the original requirements
  even if an upstream stage omitted them; explicitly report a blocker if there
  is a conflict or insufficient capability.
  \item Distinguish explicitly requested data products from the minimum inputs
  to a formula. For example, when adjusted OHLCV is requested, do not omit the
  construction, persistence, and validation of the other OHLCV fields merely
  because an EMA uses only close. Preserve all explicit parameters, windows,
  filters, rebalancing, holding, risk, and output conditions.
  \item State how each public requirement is implemented in the data pipeline,
  execution steps, or code. Do not replace unimplemented requirements with
  assumptions, defaults, or ``downstream handling.'' A textual explanation is
  not itself evidence of execution.
  \item The shared executor handles standard market execution, fees, and
  tradability constraints. It does not automatically configure strategy-specific
  selection, rebalancing frequency, holding continuation, exits, or cash branches
  from natural-language requests or strategy notes. Without an explicit
  configuration commitment, do not delegate these responsibilities to the
  Runner, backtester, or evaluator.
  \item Scores are ranking signals, not target weights. Score magnitude does
  not guarantee position size, and a non-positive score does not guarantee
  automatic exclusion. Implement the explicit selection set and rebalancing
  logic in the strategy. Do not replace weekly/monthly rebalancing with daily
  reselection or assume that omitting scores on non-rebalancing days preserves
  holdings. Distinguish retaining the previous selection from actively moving
  to cash.
  \item \texttt{downstream\_requirements} in implementation-plan and code
  outputs must list strategy requirements that still need downstream execution.
  The current generation interface does not authorize such task-specific
  delegation, so a nonempty list blocks candidate construction. If the protocol
  cannot express a requirement, report the limitation rather than claiming
  completion. Standard market matching and fees themselves do not belong in
  this list.
\end{promptitems}

\textbf{Strategy runtime interface --- candidate construction and repair.}
\begin{promptitems}
  \item The strategy class must implement only a synchronous
  \texttt{run(self, context)}; do not implement \texttt{generate\_scores} or
  \texttt{generate\_order\_signals}.
  \item \texttt{context} exposes only \texttt{dataset\_name},
  \texttt{manifest\_path}, \texttt{manifest}, and \texttt{datasets}.
  \item Read the Runner-built parquet files using
  \nolinkurl{pandas.read_parquet(context.datasets["daily"])},
  \texttt{context.datasets["minute"]}, or
  \texttt{context.datasets["snapshot"]}.
  \item Do not write \texttt{isinstance(context, dict)} or
  \texttt{context.get(...)} compatibility branches; use
  \texttt{context.datasets} directly.
  \item Do not call \texttt{get\_daily\_history},
  \texttt{get\_minute\_history}, \texttt{current\_universe}, or any legacy
  backtest-context API.
  \item \texttt{run(context)} must return
  \texttt{\{"protocol": "scores" or "orders", "data": pandas.DataFrame\}}.
  The returned protocol must match the fixed data-binding/implementation contract.
  \item Under the scores protocol, the returned columns must be
  \texttt{signal\_date}, \texttt{symbol}, and \texttt{score}, with at most one
  row per \texttt{signal\_date+symbol}.
  \item Under the orders protocol, the returned columns must include at least
  \texttt{signal\_time}, \texttt{symbol}, and \texttt{side}. Each row must
  provide one of \texttt{shares}, \texttt{notional}, or \texttt{weight}, or
  use \texttt{sell\_all=True} when selling; \texttt{side} must be
  \texttt{buy} or \texttt{sell}.
  \item Diagnostic columns prefixed with \texttt{evidence\_} may be added to
  actual decision rows to record computed condition values, thresholds, state
  changes, and results. Use separate Boolean columns for cash, exit, and
  invalidation conditions while retaining their numerical basis. These columns
  do not change trading rules or replace fill evidence. Do not add orders,
  fabricate triggers, or label an untriggered condition as verified to obtain
  coverage.
\end{promptitems}
\end{promptbox}

\paragraph{Task understanding.}
\begin{promptbox}{Prompt 1: Task understanding}
You are the Task Builder Agent of QuantHarness.

Your task is to understand the user's natural-language request and output a
structured task-understanding result. You are not a strategy-code generator:
do not write strategy code or claim that backtesting has been completed.

You need to determine:
\begin{promptitems}
  \item What type of task the user wants to perform.
  \item What the user's strategy intent is.
  \item Which key information is already explicit.
  \item Which key information is missing.
  \item Whether the user's description contains conflicts.
  \item Whether clarification is required in the current mode.
\end{promptitems}
The output must conform to the supplied structured schema.
\end{promptbox}
The model interprets the request, while parsing checks that its structured
output has the required fields and types. The task-intake policy then uses
the operating mode and recorded missing information, conflicts, and assumptions
to determine whether development may proceed or clarification is required.

\paragraph{Data binding.}
Data binding connects the planned strategy to available inputs. The agent
uses the strategy specification (StrategySpec) and planning blueprint to
select data assets, fields, and an output protocol; catalog and inspection
tools provide evidence for these choices.
\begin{promptbox}{Prompt 2: Data binding --- assets, protocol, and history}
You are the Strategy Developer Agent of QuantHarness, currently executing the
data-binding substep of the DEVELOPING stage.

Your task is to select the data assets, fields, and output protocol needed by
the strategy implementation based on the Runtime Card, StrategySpec, and
\texttt{strategy\_blueprint.md}.

You must comply with the following:
\begin{promptitems}
  \item Perform data binding only; do not write \texttt{strategy.py} or output
  executable code.
  \item First call \texttt{list\_data\_catalog} to inspect the available assets.
  \item Call \texttt{inspect\_data\_asset} for every asset to be used. Omitting
  \texttt{file\_path} returns the asset's fields, a sample parquet schema, and
  sample rows.
  \item \texttt{dataset.selections} may use only asset identifiers and fields
  confirmed to exist by \texttt{inspect\_data\_asset}.
  \item \nolinkurl{dataset.selections[*].fields} must contain source-asset field
  names, such as \texttt{close}, \texttt{amount}, and \texttt{mom\_20d}, not
  Runner output column names such as \texttt{alias\_\_field} or
  \texttt{dataset\_key.field}.
  \item The \texttt{asset\_id}, \texttt{kind}, \texttt{frequency}, \texttt{table},
  and \texttt{fields} returned by \texttt{list\_data\_catalog} are the
  highest-priority source for quant-asset namespaces. When producing
  \texttt{dataset.selections}, copy asset identifiers and fields present in the
  catalog exactly.
  \item Do not infer feature/factor namespaces from natural-language semantics.
  A ``factor'' in strategy terminology is not necessarily
  \texttt{kind="factor"} in the data system. For example, if the catalog lists
  \texttt{mom\_20d} under \texttt{quant:feature:mom}, select
  \texttt{quant:feature:mom}; do not rewrite it as \texttt{quant:factor:mom}.
  \item For any quant field, first match its name to a unique or unambiguous
  asset in the \texttt{assets} returned by \texttt{list\_data\_catalog}. If it
  appears only under \texttt{quant:feature:*}, select that feature asset; select
  a factor asset only if it appears exclusively under \texttt{quant:factor:*}.
  \item If a field cannot be confirmed in the catalog or
  \texttt{inspect\_quant\_layer} results, do not invent a quant asset or guess
  its kind/table. Record it as missing/unresolved, or select existing base-data
  fields and leave the derivation to ImplementationPlan.
  \item If the strategy involves quant features, factors, or factor ranking,
  also call \texttt{inspect\_quant\_layer}. First use \texttt{overview/table}
  to inspect available features/factors, then use \texttt{table} or
  \texttt{column} to confirm built status, output columns, coverage, and
  upstream inputs.
  \item Do not treat a quant factor with only metadata/configuration and
  \texttt{built=false} as directly readable data. Such assets must first be
  processed by external data/factor-building tools, or replaced with base/feature
  data already built in the current lake.
  \item Do not proactively call \texttt{build\_quant\_layer} during DataBinding.
  If a configuration exists but \texttt{built=false}, the Harness builds it
  after DataBinding into a run-local overlay using a controlled date window.
  \item If no ready-made local factor exists but the strategy can derive
  intermediate variables from existing base fields in \texttt{strategy.py},
  select the base-data assets and leave the derived variables to
  ImplementationPlan rather than inventing a quant asset.
  \item For daily moving-average, momentum, reversal, and factor-ranking rule
  strategies, select \texttt{protocol="scores"} by default.
  \item Select \texttt{protocol="orders"} only when the task explicitly
  requires minute-level orders.
  \item When using minute assets, set \texttt{dataset.freqs}, for example
  \texttt{["1min"]}. Do not put \texttt{freq} in \texttt{selection.fields};
  the dataset builder generates it.
  \item Do not directly join daily stock-panel fields to those of a single
  benchmark/index by \texttt{ts\_code}. When selecting
  \texttt{clean:index\_daily}, set \texttt{selection.entity\_scope="broadcast"}
  and use \texttt{selection.ts\_codes} to specify one index code whose existence
  has been confirmed by \texttt{inspect\_data\_asset}. Do not let it inherit
  the stock universe's global \texttt{ts\_codes}.
  \item Use \texttt{entity\_scope="panel"} for an ordinary stock-panel
  selection, or retain the default \texttt{auto}. Use \texttt{broadcast} only
  for a single daily entity that must be broadcast to the stock panel by date.
  \item Set \texttt{dataset.lookback\_trading\_days} according to the strategy's
  actual history requirement; a 20/60-day moving-average crossover typically
  uses 70.
  \item Check asset coverage and trading-calendar coverage before calculating
  the minimum required history. Do not convert months/years into trading days
  using minute-bar counts. Do not increase the warmup period without justification, or
  hide insufficient history by padding with zeros or shortening a publicly
  required window. If history is genuinely missing, explicitly report the
  required data range.
\end{promptitems}
The output must conform to the supplied structured schema. The top-level JSON
may contain only \texttt{protocol}, \texttt{dataset},
\texttt{field\_usage\_notes}, and \texttt{confidence}.
\end{promptbox}
The proposed binding is checked against catalog and asset-inspection evidence.
The resulting field contract (DataFieldContract) records how source fields map
to the dataset columns available to the strategy. Construction uses
this contract rather than repeating data discovery.

\paragraph{Construction and repair.}
With the binding fixed, construction turns the implementation plan into
candidate code. Prompt~3 requires this code to use the shared runtime interface:
\texttt{context.datasets} supplies input-file paths, and \texttt{run(context)}
returns the declared protocol together with a table of scores or order intents.
Protocol selection belongs to data binding; construction cannot change that
choice.
\begin{promptbox}{Prompt 3: Candidate construction --- fixed inputs and executable output}
You are the Strategy Developer Agent of QuantHarness, currently executing only
the DEVELOPING stage.

Your task is to generate a candidate strategy implementation based on the
complete public request, strategy requirements, and field contract in
\texttt{development\_context}.

You must comply with the following:
\begin{promptitems}
  \item Generate only a candidate implementation.
  \item Output the complete Python source for \texttt{strategy.py} as the
  line array \texttt{strategy\_source\_lines}, together with the minimum
  loading information: \texttt{strategy\_file}, \texttt{object\_name},
  \texttt{strategy\_kwargs}, \texttt{assumptions}, and \texttt{notes}.
  \item The Harness generates \texttt{strategy\_implementation.json} from
  StrategySpec and DataBinding. Do not output \texttt{implementation},
  \texttt{protocol}, or \texttt{dataset}.
  \item By default, use \nolinkurl{implementation/strategy.py} for
  \texttt{strategy\_file} and \texttt{Strategy} for \texttt{object\_name}.
  \item The protocol and dataset in DataBinding are immutable inputs. Do not
  reselect data or change the protocol during code generation.
  \item StrategyImplementationPlan is the auditable implementation plan prepared
  before code generation. Write code according to its \texttt{input\_datasets},
  \texttt{derived\_features}, \texttt{output\_columns},
  \texttt{execution\_steps}, and \texttt{pseudocode}.
  \item \nolinkurl{development_context.data_contract} is a compact version of
  DataFieldContract. When reading parquet files in \texttt{context.datasets},
  use only the fields declared there.
  \item Use the column names in
  \nolinkurl{development_context.data_contract.datasets[*].columns} exactly.
  Do not rewrite contract columns such as \texttt{mom\_\_mom\_20d} and
  \texttt{small\_size\_\_small\_size} into semantic aliases such as
  \texttt{momentum\_\_mom\_20d} and \texttt{size\_\_small\_size}.
  \item Do not rename strategy-ready fields to source-field names outside the
  contract and then access them by indexing. For example, do not rename
  \texttt{ret\_\_ret\_5d} to \texttt{ret\_5d} and then write
  \texttt{work["ret\_5d"]}. Use \texttt{ret\_\_ret\_5d} directly, or name
  internal derived columns clearly, such as \texttt{derived\_ret\_5d} or
  \texttt{derived\_score}.
  \item If a temporary DataFrame is created for calculation, its temporary
  columns must use internal derived names such as \texttt{derived\_*},
  \texttt{tmp\_*}, or \texttt{score\_component\_*}. Do not use bare names such
  as \texttt{close}, \texttt{amount}, \texttt{ret\_5d}, or
  \texttt{price\_position\_20d} that look like source-data fields.
  \item Rule-based strategies are the current default. If StrategySpec
  explicitly requires an ML/DL/training route, conservatively produce the
  minimal implementation that the current toolchain can audit, or explain the
  point of failure. Do not pass off simple pandas rules as actual model training.
  \item The strategy class must have a nonempty \texttt{name} attribute.
  \item Data discovery has already been completed by the DataBinding substep.
  This stage does not call data tools; write code using only DataBinding and
  DataFieldContract.
  \item Do not assume field names. Every input field must come from
  DataFieldContract.
  \item For daily moving-average, momentum, reversal, and factor-ranking rule
  strategies, select \texttt{protocol="scores"} by default; select
  \texttt{protocol="orders"} only for strategies requiring minute-level order
  matching.
  \item Tool evidence has already confirmed DataBinding's
  \texttt{dataset.selections}. Do not add or remove assets or fields at this
  stage.
  \item The Runner-built dataset parquet uses alias-prefixed column names.
  For example, selecting
  \texttt{\{"asset\_id":"clean:daily","fields":["close"]\}} without an
  alias typically produces \texttt{daily\_\_close}, not \texttt{close}.
  Prefer \texttt{daily\_\_close}; when compatibility is needed, search
  \texttt{daily.columns} for a column ending in \texttt{\_\_close} or
  containing \texttt{close}.
  \item Do not import \texttt{StrategyContext} or \texttt{OrderSignalContext}.
  \item \texttt{context.datasets} is a dictionary of parquet paths, such as
  \texttt{\{"daily": Path, "minute": Path, "snapshot": Path\}}.
  \item Missing trading days under the scores protocol are treated as empty
  scores by the backtest engine; they are not forward-filled.
  \item \texttt{strategy\_source\_lines} must be a JSON string array, with
  one line of Python source per element. Do not output a
  \texttt{strategy\_source} field or bare multiline text.
\end{promptitems}
The output must conform to the supplied structured schema. The top-level JSON
may contain only \texttt{strategy\_file}, \texttt{object\_name},
\texttt{strategy\_kwargs}, \texttt{assumptions}, \texttt{notes},
\texttt{strategy\_source\_lines}, \texttt{developer\_notes},
\texttt{downstream\_requirements}, and \texttt{confidence}.
\end{promptbox}
Candidate code is validated before backtesting. If a failure
leads to an authorized repair, Prompt~4 limits the change to the identified
engineering problem while preserving the strategy's intent. An empty output
or an untriggered condition is not, by itself, a reason to change a trading rule.
Before allowing a source change to be written back, the repair guard compares
the original and proposed code. This comparison uses their abstract syntax
trees (ASTs), which represent executable code structure.
\begin{promptbox}{Prompt 4: Repair --- preserve intent and bound the change}
You are the Strategy Repair Agent of QuantHarness, currently executing only
the REPAIRING stage.

Your task is to make the smallest-scope repair to the current candidate
strategy implementation based on public failure feedback.

You must comply with the following:
\begin{promptitems}
  \item Do not reinterpret the user's task or change the strategy intent,
  parameters, or rules defined in \texttt{strategy\_spec}.
  \item Empty output, zero trades, or an untriggered branch does not mean that
  the trading conditions are wrong. Do not relax thresholds, shorten windows,
  remove filters, change AND to OR, or add fallback entry, exit, or forced-trade
  branches to produce orders. Do not hide an exception by removing calculation
  logic and returning an empty table.
  \item Automatic write-back compares the source AST before and after repair,
  accepting only changes that preserve the logic or mechanical corrections
  explicitly supported by the validator or field contract. Preserve the
  existing code structure and avoid unrelated refactoring. Merely declaring
  ``unchanged semantics'' or retaining implementation parameters does not
  replace source-code checks.
  \item If no verifiable engineering error can be found, retain the original
  strategy and explain the untriggered conditions and data requiring inspection
  in \texttt{root\_cause}, \texttt{intended\_fix}, and
  \texttt{developer\_notes}. Do not turn a no-trade strategy into one that
  trades more easily, or claim that it has been repaired.
  \item By default, modify only
  \nolinkurl{candidate/implementation/strategy.py}; do not change
  \texttt{strategy\_file}, \texttt{object\_name}, or \texttt{protocol}.
  \item Do not remove data assets, fields, or existing
  \texttt{strategy\_kwargs} parameters declared in
  \texttt{current\_implementation}. If passing requires deleting assets or
  fields, changing the protocol, or changing existing parameters, retain these
  contracts and explain in \texttt{repair\_plan}, \texttt{root\_cause}, and
  \texttt{developer\_notes} that human or data-side repair is needed.
  \item You may extend an explicitly specified
  \texttt{lookback\_trading\_days} or add necessary fields. Do not shorten
  the history window or change \texttt{as\_of\_date}, \texttt{freqs}, or
  the stock universe to obtain output. If these data boundaries must change,
  retain the current candidate and report the data-side problem.
  \item If the failure comes from \texttt{validation\_report}, repair only the
  engineering/interface problems identified in that report.
  \item If the failure comes from \texttt{backtest\_report}, repair only the
  runtime problems exposed by the traceback or error message.
  \item If \texttt{where(..., shares.astype(int))} raises
  \texttt{IntCastingNaNError} because a \texttt{sell\_all} row has no quantity,
  convert only non-null rows:
  \nolinkurl{orders.loc[orders["shares"].notna(), "shares"] =}
  \nolinkurl{orders["shares"].dropna().astype(int)}.
  Preserve missing quantities and all trading conditions. Do not hide the error
  with \texttt{fillna(0)} or by deleting sell orders.
  \item Retain the strategy class's \texttt{name}; do not add dangerous imports
  or calls.
  \item If the failed validation check is
  \texttt{strategy\_name\_consistency}, align
  \texttt{implementation.strategy\_name} only to
  \texttt{strategy\_spec.strategy\_name}; do not change it to a third name.
  \item If \texttt{failure\_feedback.repair\_mode} is \texttt{code\_only}, this
  round may repair only \texttt{strategy\_source}. Copy
  \texttt{strategy\_name}, \texttt{protocol}, \texttt{strategy\_file},
  \texttt{object\_name}, \texttt{strategy\_kwargs}, and \texttt{dataset} in
  \texttt{implementation} unchanged from \texttt{current\_implementation}.
  \item \texttt{strategy\_source} must be a plain Python source-code string,
  not wrapped in a Markdown code block.
\end{promptitems}
The output must conform to the supplied structured schema. The top-level JSON
may contain only \texttt{repair\_plan}, \texttt{implementation},
\texttt{strategy\_source}, \texttt{developer\_notes}, and \texttt{confidence}.
\end{promptbox}
Alongside source checks, the repair guard checks changes to the implementation
fields; code-only repair retains the non-code fields. The revised candidate
must then pass validation again, including checks of its
interface, data access, and executable output. Data-side problems requiring
broader changes are reported or routed through separately authorized recovery,
not silently solved by changing the evaluation universe or dates.

\paragraph{Judge review.}
The internal Judge reviews the candidate using validation and backtest
evidence. It supports rule-based and configured LLM-assisted modes;
Prompt~5 applies only to the latter. Its compact evidence input summarizes the
task, candidate, and execution reports, while \texttt{artifact\_references}
lists the available supporting artifacts.
\begin{promptbox}{Prompt 5: LLM Judge --- evidence review}
You are the LLM Judge of QuantHarness.

LLM Judge evidence boundary:
\begin{promptitems}
  \item Perform semantic review using only public compact evidence.
  \item Do not invent evidence absent from \texttt{artifact\_references}.
  \item Do not bypass hard validation or backtest failures.
  \item \texttt{evidence\_paths} may refer only to paths in
  \texttt{artifact\_references} with \texttt{exists=true}.
  \item Focus on whether the strategy matches the user's intent, whether the
  backtest evidence is sufficient, and whether no trades or few trades are more
  indicative of an implementation error, an insufficient window, data
  limitations, or a legitimate no-position state.
\end{promptitems}
\end{promptbox}
The Judge guard checks the LLM's cited evidence paths and whether its
recommendation is compatible with hard validation and backtest prerequisites.
The recommendation does not itself authorize freezing or a state transition.
Tool permissions, permitted state transitions, and budget accounting are
enforced by the runtime, as are candidate-identity checks when seal enforcement
is enabled (Appendix~\ref{app:stage-controls}). These are deterministic checks,
not additional language-model judgments.
This internal review is distinct from the independent requirement-level panel
in Appendix~\ref{app:semantic-panel}.
\FloatBarrier

\subsection{Market and Backtest Protocol}
\label{app:market-protocol}

All reported financial outcomes are recomputed by one deterministic A-share
evaluator rather than taken from a system's native backtester. This keeps the
price basis, transaction costs, liquidity limits, and market rules fixed within
each comparison.

The FinRelBench snapshot uses the CSI~500 constituent list dated 2025-12-31,
with daily coverage from 2025-01-02 through 2026-05-29 and minute coverage from
2026-04-29 through 2026-05-29. The provider is Tushare; snapshot manifests record
checksums that identify the exported clean and quantitative data files.
Individual tasks request the history they need within this coverage. Daily
history before the evaluation window provides warmup observations for computing
the task's signals. When a task requests adjusted prices, evaluation inspects
both the adjustment-factor inputs and the code that computes those prices.
The separate boundary tests use a frozen 100-name CSI~100 universe for checking
whether a candidate passes backtesting.

FinResearchSuite's chronological windows and CSI~500 constituent snapshots
are specified in Appendix~\ref{app:finresearchsuite-protocol}.

\subsubsection{Signal Timing, Portfolio Conversion, and Execution Constraints}

Daily scores generated at market close cannot trade on that close. They are
converted to orders for the next session, using the 09:30--10:00
volume-weighted average price for execution. Unless a public task specifies
another construction rule, daily-score outputs form equal-weight, long-only
portfolios with at most 100\% gross exposure.

Intraday order outputs retain their declared shares, notional amounts, or
weights. Execution is attempted on the first available minute bar at least
one minute after the signal time, using the price basis stated by the task.
Both timing rules prevent execution on the bar that produced the signal and
apply equally to systems within a case. Actual fills remain subject to the
account and market constraints below.

Each account starts with CNY~1,000,000. The evaluator charges a three-basis-point
brokerage commission with a CNY~5 minimum, a five-basis-point stamp duty on
sales, and a 0.1-basis-point transfer fee on both purchases and sales. The
experimental default slippage is five basis points; the filtered-reversal case
has a prespecified eight-basis-point override.

Beyond trading costs, fills are constrained by liquidity, available cash, and
market rules. Each order may consume at most 10\% of market volume in its
execution window or minute bar. The account prohibits negative cash and does
not allow purchased shares to be sold until the next trading session (T+1).
It also enforces trading suspensions, daily price limits, and purchases in
multiples of 100 shares, while allowing complete liquidation of an odd-lot
remainder.

Case-specific lookback, portfolio, and execution constraints are fixed across
systems within a task. These settings define the common evaluation conditions;
they enter requirement-fidelity assessment only when specified in the public
request.

\subsection{Systems, Budgets, and Environments}
\label{app:systems-environments}

This subsection records the software, generation budgets, and independent
evaluation environment used for FinRelBench. Strategy-generation calls use
GPT-5.5 at temperature 0.0 where the runtime exposes that parameter.
QuantHarness accesses the frozen market data through its governed data tools;
external systems receive the corresponding read-only files or catalog views
through their native runtime.

\paragraph{Generation budgets.}
Table~\ref{tab:system-freeze} records the system releases and native generation
limits used in this comparison. These native limits operate within a shared,
task-dependent time limit for the whole generation process, specified in
Table~\ref{tab:reliability-budgets}. Thus, the comparison bounds elapsed time
without imposing a common iteration or tool-use policy on all systems.

\begin{table*}[t]
  \caption{FinRelBench system releases and native generation limits.}
  \label{tab:system-freeze}
  \centering
  \small
  \setlength{\tabcolsep}{6pt}
  \begin{tabular*}{\textwidth}{@{\extracolsep{\fill}}lll@{}}
    \toprule
    System & Implementation / release & Native limit within the process cap \\
    \midrule
    QuantHarness & Native Workflow & 20 turns; one repair allowance \\
    Codex CLI & \texttt{exec} 0.144.1 & No comparable turn cap exposed \\
    OpenHands & agent-server 1.8.0 (\texttt{de4e2eb}) & 80 native iterations \\
    SWE-agent & runner 1.1.0 (\texttt{abd7d69}) & 2.00 per-instance cost limit \\
    Aider & CLI 0.86.3.dev53 (\texttt{5dc9490}) & No comparable turn cap exposed \\
    \bottomrule
  \end{tabular*}
\end{table*}
\FloatBarrier

For QuantHarness, the turn limit applies to each agent invocation; the whole
generation-process time limit bounds the task across stages. Native limits
and the shared time limits are separate controls, not interchangeable units.
Table~\ref{tab:reliability-budgets} distinguishes the limits for generation
from those for independent dataset construction and strategy execution.

\begin{table}[htbp]
  \caption{FinRelBench time limits in seconds, separating strategy generation
  from independent evaluation for each output type.}
  \label{tab:reliability-budgets}
  \centering\small
  \begin{tabular}{lrr}
    \toprule
    Operation & Daily-score task & Intraday-order task \\
    \midrule
    Whole generation process & 1,200 & 2,400 \\
    Individual model request & 300 & 600 \\
    Agent stage, including request retry & 660 & 1,260 \\
    Data-binding stage & 660 & 1,260 \\
    \midrule
    Independent dataset construction & 300 & 600 \\
    Independent strategy execution & 150 & 300 \\
    \bottomrule
  \end{tabular}
\end{table}

The model-request policy permits one retry. Stage limits apply where the
native runtime exposes the corresponding stage. Infrastructure recovery
addresses service interruptions separately from native strategy repair.
The comparison reports these configured budgets and observed outcomes;
product-specific token and tool counters are not treated as a common
efficiency metric.

\paragraph{Independent evaluation.}
Declared data-construction and strategy-execution commands run in the container
image \texttt{quantharness-contract-neutral:py311-v1} through the common
submission interface. External panel calls use temperature 0.0, a 600-second
request timeout, a maximum of three runtime turns, and two infrastructure
retries. The panel procedure and evidence collection are described in
Appendices~\ref{app:semantic-panel} and~\ref{app:benchmark-evidence}.

\paragraph{Reproduction requirements.}
\label{app:artifact-reproduction}
Recomputing aggregate results or reviewing saved evidence does not require
new strategy-generation calls. Execution and decision replay require complete
retained inputs and matching runtime configurations
(Appendix~\ref{app:audit-replay-protocol}); strategy execution also requires
authorized access to the same market-data snapshot. Regenerating a development
trajectory requires new model calls, access to the corresponding model, the
complete prompts and tool definitions used for the run, and matching runtime
settings. Appendix~\ref{app:prompt-constraints} presents core instruction
blocks rather than complete run-specific requests.

\end{document}